\documentclass{article}

\usepackage{arxiv}
\usepackage[utf8]{inputenc}
\usepackage{xcolor}
\usepackage{graphicx}
\usepackage[square,numbers,sort]{natbib}
\usepackage{color}
\usepackage{soul}
\usepackage{siunitx}
\usepackage[hidelinks,breaklinks=true]{hyperref}
\usepackage{booktabs}
\usepackage{longtable}
\usepackage{tabularx}
\usepackage{endnotes}
\usepackage[defaultcolor=red]{changes}

\usepackage{caption} 


\let\footnote=\endnote

\title{Review of Natural Language Processing in Pharmacology}

\author{
  \textbf{Dimitar Trajanov$^1$ $^2$, Vangel Trajkovski$^1$, Makedonka Dimitrieva$^1$, Jovana Dobreva$^1$,} \\ \textbf{Milos Jovanovik$^1$, Matej Klemen$^3$, Aleš Žagar$^3$,  Marko Robnik-Šikonja$^3$} \\
  $^1$ Ss. Cyril and Methodius University in Skopje, Faculty of Computer Science and Engineering, North Macedonia \\
  $^2$ Boston University, Computer Science Department, Metropolitan College,  Boston, MA, USA \\
  \texttt{\{dimitar.trajanov,jovana.dobreva,milos.jovanovik\}@finki.ukim.mk} \\
  \texttt{\{vangel.trajkovski,makedonka.dimitrieva\}@students.finki.ukim.mk} \\
  $^3$ University of Ljubljana, Faculty of Computer and Information Science, Slovenia \\
  \texttt{\{matej.klemen,ales.zagar,marko.robnik\}@fri.uni-lj.si}
}

\begin{document}
\maketitle

\begin{abstract}
     Natural language processing (NLP) is an area of artificial intelligence that applies information technologies to process the human language, understand it to a certain degree, and use it in various applications. This area has rapidly developed in the last few years and now employs modern variants of deep neural networks to extract relevant patterns from large text corpora. The main objective of this work is to survey the recent use of NLP in the field of pharmacology.
     As our work shows, NLP is a highly relevant information extraction and processing approach for pharmacology. It has been used extensively, from intelligent searches through thousands of medical documents to finding traces of adversarial drug interactions in social media. We split our coverage into five categories to survey modern NLP methodology, commonly addressed tasks, relevant textual data, knowledge bases, and useful programming libraries. We split each of the five categories into appropriate subcategories, describe their main properties and ideas, and summarize them in a tabular form. The resulting survey presents a comprehensive overview of the area, useful to practitioners and interested observers.
\end{abstract}

\section{Introduction}
Information processing is indispensable to modern drug design, production, and application. A significant amount of information is stored in textual format and located in scientific papers, clinical notes, ontologies, knowledge bases, social media posts, and newspaper articles. Extraction and retrieval of this information rely on natural language processing (NLP). NLP is a broad scientific area based on computer science, linguistics, and artificial intelligence \citep{jurafsky2008speech,jurafsky2022speech}. As the whole area of artificial intelligence, it has been completely transformed in recent years by deep learning \citep{Goodfellow2016}. It has witnessed numerous new techniques and successful applications, such as intelligent search, machine translation, and speech recognition. 

Many general NLP techniques and approaches can be applied to the pharmacological area. However, often NLP techniques have to be adapted to the specifics of the field in terms of available knowledge sources, text representation, specific methods, terminology, etc. 
In this work, we survey modern NLP methodology, tasks, resources, knowledge bases, and tools used and adapted to the area of pharmacology. The review aims to inform practitioners working in the area of the pharmacology of exciting recent development and to give a solid starting reference material to new entrants.

Several surveys summarise NLP in pharmacology but only cover specific areas of NLP methods. One of the first reviews of NLP for clinical decision support (CDS) \citep{demner2009can} was published in 2009. The authors observed that many CDS data is textual and reviewed existing NLP developments for CDS. \citet{luo2017natural} present a structured review of NLP for narratives in electronic health records (EHR) for pharmacovigilance.  \citet{dreisbach2019systematic} review NLP of symptoms from electronic patient-authored text data. A review of NLP in languages other than English for clinic-related texts is presented by \citet{neveol2018clinical}. \citet{chen2021artificial} survey NLP  addressing challenges related to COVID-19 pandemic. They present details related to several NLP tasks like information retrieval, named entity recognition, literature-based discovery, question answering, topic modeling, sentiment and emotion analysis, caseload forecasting, and misinformation detection. In contrast to the listed surveys, we aim for a comprehensive overview of NLP in pharmacology. 

NLP is a subfield of much broader areas of machine learning (ML) and artificial intelligence (AI). Many ML algorithms not related to text are applicable to pharmacological tasks. While the focus of this paper is the application of NLP in pharmacology, we here refer to several recent survey articles that cover ML application in specific pharmacological tasks, like drug discovery \citep{stephenson2019survey,dara2022machine, carracedo2021review}, drug-target interaction prediction \citep{le2016systems, chen2018machine}, drug repurposing \citep{yang2022machine}, drug-drug interactions \citep{han2022review}, ML applications for COVID-19 \citep{kamalov2021machine}, pharmacometrics \citep{janssen2022adoption,mccomb2022machine}, cancer management \citep{kumar2022review}, microbiome therapeutics \citep{mccoubrey2021harnessing}, exploratory pharmacovigilance \citep{kaas2022exploratory}, biomaterials \citep{kerner2021machine}, and many more.

Recently, the primary methodological approach to NLP has been deep learning. Deep neural networks (DNNs) require that text is transformed (embedded) into  numeric vectors in a process called representation learning. We present general text embeddings as well as specific variants relevant to the area of life sciences and pharmacology. As pharmacology is a knowledge-intensive area where relevant information is not stored only in text documents but also in databases, ontologies, and linked data, we survey recent attempts to inject knowledge into DNNs. Due to the need to understand the decisions and biases of DNNs, we discuss techniques that make their output more transparent. 

Some NLP tasks are particularly important to the area of pharmacology. While they are often based on general approaches, they are strongly adapted and use specific pharmacological resources. We discuss general tasks such as named entity recognition, relation extraction, literature-based discovery, question answering, and field-specific tasks such as detection of adverse drug reactions and drug discovery.

The basic precondition for applying NLP is the availability of language resources. In multiple studies, EHRs are the main source of information \citep{wunnava2019adverse,liu2019towards,wang2009active,jagannatha2019overview,li2018extraction}. EHRs contain patient data such as diagnoses, hospital admissions, prescriptions, and adversary drug effects. The data in EHRs are well-structured and can be readily processed; however, different EHR components are difficult to integrate. Many authors use molecular data \citep{suthram2010network,park2011protein}, which can be integrated with diseases \citep{goh2007human}.  Other important sources of information are clinical data \citep{jung2013inferring} (used in, e.g., drug repurposing \citep{yang2017literature, deftereos2011drug}), linked data, and the pharmacology-related semantic web. 

Linked data and knowledge graphs have recently emerged as general formalisms to represent knowledge in artificial intelligence and the semantic web. Linked (open) data movement introduced new standards for representing, storing, and retrieving data over the web \citep{bizer2008linked,bizer2009linked,heath2011linked,wood2014linked,10.1145/3447772}, which enabled new distributed data sources and new applications. Knowledge graphs allow generating, consolidating, and contextually linking structured data. We present several knowledge graphs from the biomedical domain and outline several COVID-19-related knowledge graphs. 

Software tools and libraries are essential for using NLP in pharmacological research and practice. Mostly, these support the Python language. We present many general NLP tools and libraries as well as life-science and pharmacology-specific variants. 

We organize the survey along with five main areas: methodology, common tasks, datasets, knowledge graphs, and software libraries. In Section \ref{sec:NLPmethodologies}, we structure NLP methodologies into three groups: representation learning (i.e., different embeddings), approaches to inject domain-specific knowledge into deep neural networks, and explainable AI techniques used in pharmacology. The most frequently used NLP tasks in pharmacology are presented in Section \ref{sec:NLPtasks}. We cover the named entity recognition, relation extraction, adverse drug reactions, literature-based discovery, and question answering. In Section \ref{sec:data}, we first outline the approaches to finding data resources, followed by a survey of existing data. We organize the overview into five categories: patient data, drug usage data, drug structure data, question answering datasets, and general text processing datasets. Knowledge graphs used in the biomedical domain and a specific example of COVID-19 disease knowledge graphs are covered in Section \ref{sec:knowledgegraphs}. We give an overview of useful NLP software libraries and tools for the pharmacological domain as well as useful general NLP libraries in Section \ref{sec:tools}. We conclude the survey in Section \ref{sec:conclusions}.

\section{NLP Methodology in Pharmacology}
\label{sec:NLPmethodologies}

Recently, NLP has switched entirely to deep neural networks, mostly large language models (LLMs) that are pretrained on huge quantities of text to capture various linguistic, general, and domain-specific knowledge. LLMs embed the text data into a numeric representation preserving semantic relations between words. To be used for specific tasks, LLMs are fine-tuned with problem-specific data. 

In Section \ref{sec:representation}, we give an overview of modern text representations. We present static and contextual embeddings (i.e., LLMs) and specific variants relevant to the area of life sciences and pharmacology. While most of the work is focused on English, we present some notable exceptions in other languages. As pharmacology is a knowledge-intensive area where relevant information is not stored only in text documents but also in databases, ontologies, and linked-data, we survey recent attempts to inject knowledge into deep neural networks in Section \ref{sec:knowledgeInjection}. Unfortunately, deep neural networks often appear as black-box models, lacking transparency on how the decisions are taken. In Section \ref{sec:explanation}, we present general explanation techniques applicable to text prediction and focus on successful applications related to pharmacology.

\subsection{Representation Learning}
\label{sec:representation}

In NLP, text representation is a crucial issue and research direction. Various text embeddings emerged that capture both syntax and semantics of a given text. While traditional approaches were based on sparse representations such as bag-of-words, dense representations such as word2vec \citep{Mikolov2013distributed}, ELMo \citep{Peters2018}, and BERT \citep{devlin2019bert} are based on neural networks and offer much more semantically valid and computationally efficient representations. 
A common trait of these embeddings is to train a neural network on self-supervised text classification tasks and use the weights of the trained neural network or the whole trained network to represent different text units (words, sentences, or documents). 
The labels required for training these classifiers originate from large corpora of general texts, e.g., web crawl, news, and Wikipedia. The usual classification tasks used in training these representation models are predicting the next and previous word in a sequence or filling in missing words (also called masked language modeling). Representation learning can be extended with other related tasks, such as prediction if two sentences are sequential. The positive instances for learning are obtained from the text in the given corpus, while the negative instances are mostly sampled from instances that are unlikely to be related.
 
We first briefly describe the principle of the most frequently used static embeddings, called word2vec, followed by large language models such as contextual BERT. Next, we cover the adaptations of these representation techniques for life sciences and pharmacology domains. We provide a summary of the presented embeddings at the end of the section in Table \ref{tab:embeddings}.

\subsubsection{Static Embeddings}
 \label{sec:word2vec}
 
The word2vec word embedding method \citep{Mikolov2013distributed} trains a shallow (one hidden layer) neural network predicting the neighboring words of a given input word. The trained weights of the hidden layer produce a static embedding in the sense that we get a single vector for each word. For example, the term \emph{bank} may denote a financial institution or land alongside a river, but it is represented with a single vector. 

The word2vec method pre-trains a feed-forward neural network on a huge corpus, and the weights of the hidden layer in this network are used as word embeddings. Pretrained word vectors for many languages are publicly available. The published vectors are typically 100 or 300-dimensional, e.g., Google published vectors for 3 million English words and phrases (\url{https://code.google.com/archive/p/word2vec/}). While the word2vec algorithm consists of two related methods, we describe only the skip-gram method, which mostly produces more favorable results. The method constructs a neural network to classify cooccurring words by taking a word and predicting its $d$ preceding and succeeding words, e.g., $\pm ~ 5$ words. In the actual neural network, one word is on the input (the central word) and one word is on the output, where both are represented with one-hot encoding. The words and their contexts appearing in the training corpus constitute the training instances of the classification problem. The first word of the training pair is presented at the network's input in the one-hot-encoding representation, and the network is trained to predict the second word. The difference in prediction is evaluated using a loss function. For a sequence of $T$ training words $w_1, w_2, w_3, \dots ,w_T$ , the skip-gram model maximizes the average log probability:
$$
\frac{1}{T} \sum_{t=1}^T \sum_{-d \leq j \leq d, j \neq 0} \log{ p(w_{t+j}|w_t)}.
$$

Once the network is trained with word2vec, vectors for each word in the vocabulary can be generated. As one-hot encoding of the input word only activates one input connection for each hidden layer neuron, the weights on these connections constitute the embedding vector for the given input word. 

The resulting word embeddings' properties depend on the context's size. For a small number of neighboring words (e.g., $\pm ~ 5$ words), we get embeddings that perform better on syntactic tasks. For larger neighborhoods (e.g., $\pm ~ 10$ words), the embeddings better express semantic properties.

Word2vec has attracted the immense attention of NLP researchers and practitioners. The word2vec precomputed embeddings soon became a default choice for the first layer of many classification deep neural networks. Several domain-specific variants have also been created and made publicly available. For life sciences, a well-known example is the work of \citet{pyysalo2013distributional}, who released two sets of word2vec vectors. The first, denoted PubMed-PMC, was trained on 23M PubMed abstracts and 0.7M PubMed Central (PMC) articles. The second model, Wiki-PubMed-PMC, was prepared using the same two corpora combined with 4M English Wikipedia articles. These static embeddings were successfully used in many life-science applications (the paper received 492 citations by 13 March 2022). For example, \citet{Habibi2017deep} have successfully applied the two embeddings to the biomedical named entity recognition problem to detect genes, chemicals and diseases. 

Note that the same technology to represent text can be applied to represent biological sequences, such as DNA, RNA, and proteins \citep{asgari2015continuous}. The created bio-vectors (BioVec) refer to biological sequences in general, protein-vectors are called ProtVec, and gene vectors are named GeneVec. A similar attempt to represent biological sequences is dna2vec vectors \citep{ng2017dna2vec}.

Despite the successful use of static embeddings such as word2vec, contextual embedding models such as BERT have become even more successful. Therefore, we skip the detailed review of static embedding models and focus on contextual models.

\subsubsection{Contextual Word Embeddings}
\label{sec:contextualEmbeddings}

The problem with word2vec embeddings is their failure to express polysemous words. During its training, all senses of a given word (e.g., \emph{paper} as a material, as a newspaper, as a scientific work, and as an exam) contribute relevant neighboring words in proportion to their frequency in the training corpus. This causes the final vector to be placed somewhere in the weighted middle of all words' meanings. Consequently, rare meanings of words are poorly expressed with word2vec, and the resulting vectors do not offer good semantic representations. For example, none of the 50 closest vectors of the word \emph{paper} is related to science.

The idea of contextual word embeddings is to generate a different vector for each word's context. The context is typically defined sentence-wise. This solves the problems with word polysemy. The context of a sentence is mostly enough to disambiguate different meanings of a word for humans and learning algorithms. Several contextual embeddings have been developed, e.g., ELMo, ULMFit, and BERT. As the latter achieves the best results in most NLP tasks, we describe it below. 

Contextual embeddings are based on the idea of language models, which predict either the next, previous or missing word in a sequence. Training often combines several of these and other related tasks. Due to the network's depth, extracting vector representations from the network is no longer trivial, i.e., the trained deep networks store their knowledge in weights spread over several layers. A frequently used approach concatenates weights from several layers into a vector. Still, often it is more convenient to use the whole pretrained neural language model as a starting point and fine-tune its weights further during the training on a specific task. 

BERT (Bidirectional Encoder Representations from Transformers) embeddings \citep{devlin2019bert} generalize the idea of language models (LMs) to masked language models, inspired by the gap-filling tests. The masked language model randomly masks some of the tokens from the input. The task of an LM is then to predict each missing token based on its neighborhood. BERT  uses the transformer architecture of neural networks \citep{Vaswani2017} in a bidirectional sense (forward and backward). It introduces another task of predicting whether two sentences appear in a sequence. The input representation of BERT is sequences of tokens representing sub-word units. The input is constructed by summing the corresponding token, segment, and position embeddings.

Using BERT for classification requires adding connections between its last hidden layer and new neurons corresponding to the number of classes in the intended task. The fine-tuning process is typically applied to the whole network. All the BERT parameters and new class-specific weights are fine-tuned jointly to maximize the log-probability of the correct labels.

BERT has shown excellent performance on many NLP tasks and is now a de-facto standard in NLP. In the initial evaluation \citep{devlin2019bert}, BERT showed improved performance on all eight tasks from the GLUE (general language understanding evaluation) benchmark suite \citep{Wang2018}, consisting of question answering, named entity recognition, and common-sense inference. A variant of BERT, called RoBERTa \citep{liu2019roberta}, which only uses masked language model training but on a larger dataset and for a longer time, has become a popular practical choice due to its improved robustness and better parallel training capability.

Due to its success, BERT has spurred an immense tide of research, analyzing its capabilities and using and adapting it for different purposes. An overview of research on BERT capabilities and inner workings is presented by \citet{rogers2020primer}. Below, we overview the adaptations and applications relevant to pharmacology. 

\subsubsection{BERT Variants Relevant to Pharmacology}
\label{sec:BERTs}

BERT has many extensions in architecture, training, and fine-tuning. A general improvement for science-related text processing is \textbf{SciBERT} \citep{beltagy-etal-2019-scibert} that was trained on 1.14M scientific papers (3.17B tokens) from Semantic Scholar instead of general text. The training data consisted of 18\%  computer science papers and 82\% papers from the biomedical domain. Upon its introduction, the SciBERT was compared to BERT and achieved improved performance in a study involving four classification tasks based on scientific publications: named entity recognition (NER), extraction of participants, interventions, comparisons, and outcomes in clinical trial papers, text classification, relation classification, and dependency parsing (DP). The SciBERT has attracted considerable attention of the scientific community with more than \num{1000} citations recorded by Google Scholar at the time of this writing. 

In life sciences, there are several popular domain adaptations of BERT. \textbf{BioMed-RoBERTa}-base \citep{gururangan2020don} (almost \num{600} Google Scholar citations at the time of this writing) is an adaptation of RoBERTa \citep{liu2019roberta}, using long pretraining on 160GB of standard texts and additional 47GB (7.55B tokens from 2.68M papers) of abstracts and full papers randomly sampled from PubMed repository. Using this domain-adapted pretrained model, the authors improved classification for two domain-specific tasks. First, they improved the classification compared to the baseline RoBERTa model for 2.3 micro $F_1$ percent on the Chem-Prot database  \citep{kringelum2016chemprot} that contains chemical-protein-disease annotations enabling the study of systems pharmacology for a small molecule across multiple layers of complexity from molecular to clinical levels. Second, they tested the BioMed-RoBERTa on the PubMed sequential sentence classification task \citep{dernoncourt-lee-2017-pubmed} and achieved 0.4 micro $F_1$ percent advantage over RoBERTa.

The \textbf{BioBERT} \citep{Lee2019BioBERT} representation model (almost \num{2000} Google Scholar citations at the time of this writing) was initialized with BERT weights and then pretrained using domain-specific literature, namely PubMed abstracts (4.5B words) and PubMed Central full-text articles (13.5B words). The resulting model was successfully fine-tuned for three biomedical text mining tasks: biomedical named entity recognition, biomedical relation extraction, and biomedical question answering. The BioBERT model was further pretrained for clinical texts using 2M generic clinical notes and discharge summaries \citep{alsentzer-etal-2019-publicly}. The resulting \textbf{Bio+Clinical BERT} showed superior results on clinical NER tasks and medical natural language inference task.

\textbf{Clinical BERT} \citep{huang2019clinicalbert} is similar to the above Bio+Clinical BERT model, but it is trained on \num{2083180} anonymized clinical notes from the MIMIC III database \citep{johnson2016mimic} that consists of the electronic health records of \num{58976} unique hospital admissions from \num{38597} patients in the intensive care unit between 2001 and 2012. The model performed better than BERT on the clinical readmission prediction problem. A similar model is \textbf{BLUE BERT} \citep{peng2019transfer}, trained on more than 4B PubMed abstracts and 500M MIMIC-III  clinical notes. The model showed good performance on BLUE (Biomedical Language Understanding Evaluation) benchmark that includes several tasks relevant to pharmacology, like named entity recognition (see Section \ref{sec:ner}) and relation extraction (see Section \ref{sec:re}).

In the light of COVID-19 epidemics, \citet{khadhraoui2022survey} have prepared a specialized BERT model, called \textbf{CovBERT}, intended to improve the COVID-19 literature review. The model, based on BERT, was pretrained on  \num{4304} PubMed abstracts on several topics such as COVID-19 treatment, COVID-19 symptoms, virology, public health, and mental health. CovBERT showed better classification accuracy on this dataset compared to baseline RoBERTa, AlBERT, SciBERT, BioBERT, and Bio+Clinical BERT.

Another popular adaptation to specific terminological areas is named \textbf{CharacterBERT} \citep{el2020characterbert}. Instead of using subword tokenization, this approach starts with characters and first constructs words with a convolutional neural network. The pretraining used around 1B tokens from the MIMIC-III clinical dataset and PubMed abstracts. The effectiveness of this approach was originally demonstrated in the biomedical domain using four tasks: medical entity recognition, medical natural language inference, relation extraction (Chem-Prot database and drug-drug interactions), and clinical sentence similarity. The resulting CharacterBERT models performed on par or better than BERT.

As evident from many citations, the BERT enhancements received, these models were successfully applied to many relevant pharmacological problems. We list a sample of works addressing a few relevant problems and approaches in Section \ref{sec:NLPtasks}.

\subsubsection{Languages Other than English}

While the majority of NLP in pharmacology is focused on English, there are also some exceptions. \citet{Akhtyamova2020} trains a domain-specific BERT model for Spanish on a relatively small dataset (87M tokens) and successfully applies it to the problem of NER in Spanish. In the context of the annual workshop on BioNLP Open Shared Tasks, in 2019 (\url{https://2019.bionlp-ost.org/}) one of the tasks, PharmaCoNER (Pharmacological Substances, Compounds and proteins and Named Entity Recognition track),  addressed the mentioning of chemicals and drugs in Spanish medical texts. The task included two tracks: one for the NER offset and entity classification and the other one for the concept indexing. In their entry, \citet{xiong-etal-2019-deep} devised a system based on BERT for the NER offset and entity classification and Bi-LSTM with max/mean pooling for concept indexing. On the same tasks, \citet{sun2021deep} compared several BERT variants (see Section \ref{sec:BERTs}): BLUE BERT \citep{peng2019transfer}, multilingual BERT \citep{devlin2019bert}, SciBERT \citep{beltagy-etal-2019-scibert}, BioBERT \citep{Lee2019BioBERT}, and Spanish BERT \citep{canete2020spanish}. The results show that domain-specific pretraining is successful and better than the language-specific BERT variant.

For the adverse drug reaction relation extraction in Russian, \citet{Sboev2022} have preliminary trained multilingual XLM-RoBERTa \citep{conneau2020unsupervised}, and Russian RuBERT \citep{kuratov2019adaptation} models on Russian drug review texts, followed by fine-tuning on the created training dataset. The results showed that the former multilingual model is advantageous. \citet{Tutubalina2020} have created a consumer reviews corpus in Russian about pharmaceutical products for the detection of health-related named entities and the assessment of pharmaceutical product effectiveness. Using this corpus and the multilingual BERT they created domain specific RuDR-BERT which showed favorable performance on medical named entity recognition and multilabel sentence classification.

\begin{longtable}[l]{p{2.2cm}|p{3.6cm}|p{4.2cm}|p{4.7cm}}
        \toprule
        \textbf{Name} & \textbf{Description} & \textbf{Trained on} & \textbf{Usage}\\
        \toprule
        \multicolumn{4}{c}{\textbf{Static embeddings}}\\
        \midrule
        \endfirsthead
        word2vec \citep{Mikolov2013distributed} & General static word embeddings & Any collection of text, e.g., Wikipedia dump & Any general non-contextual text processing.\\
        \midrule
        PubMed-PMC, WikiPubMed-PMC \citep{pyysalo2013distributional} & Word2vec adapted to life-sciences & PubMed abstracts and articles; in combination with Wikipedia & Any non-contextual life-science text processing, e.g., biomedical NER for genes, chemicals and diseases \citep{Habibi2017deep}\\
        \midrule
        BioVec, ProtVec, GeneVec  \citep{asgari2015continuous} 
        dna2vec \citep{ng2017dna2vec} & Word2vec style embeddings for biological sequences, genes, and proteins & Different biological sequences, e.g., Swiss-Prot & Proteomics and genomics, e.g., structure prediction for proteins. \\
        \midrule
        \multicolumn{4}{c}{\textbf{Contextual embeddings}}\\
        \midrule
        BERT \citep{devlin2019bert}, RoBERTa \citep{liu2019roberta} & General contextual text embeddings & Large general text corpora such as Wikipedia and Common Crawl. & Any general text processing.\\
        \midrule
        SciBERT \citep{beltagy-etal-2019-scibert}  & Contextual embeddings for scientific texts. & Scientific papers from Semantic Scholar & NER for clinical use, text classification, relation classification \\
        \midrule
        Character BERT \citep{el2020characterbert} & Character-level input allows for easy adaptation to different areas. & Clinical texts and PubMed abstracts. & Medical NER, NLI, RE, and  clinical sentence similarity. \\
        \midrule
        BioMed-RoBERTa \citep{gururangan2020don} & RoBERTa adaptation for life-sciences & Standard texts,  abstracts, and full papers from PubMed. & Chemical-protein-disease annotations, sequential sentence classification task \\
          \midrule
        BioBERT \citep{Lee2019BioBERT} & BERT adapted to life-sciences & BERT further trained on PubMed abstracts and papers & biomedical NER, RE, and QA \\
        \midrule
         Bio+Clinical BERT \citep{alsentzer-etal-2019-publicly} & BioBERT adapted to clinical texts & BioBERT  further pretrained with clinical notes and discharge summaries. & Clinical NER and medical NLI. \\
           \midrule
        Clinical BERT \citep{huang2019clinicalbert}  & Suitable for clinical texts & Clinical notes from EHR for patients in intensive care units. & Clinical readmission prediction. \\
          \midrule  
        BLUE BERT \citep{peng2019transfer} & Suitable for clinical texts & PubMed abstracts and clinical notes. & Good performance on BLUE benchmark, including NER and RE. \\
            \midrule
        CovBERT  \citep{khadhraoui2022survey} & BERT adapted to COVID-19. &  BERT further pretrained  on PubMed abstracts with COVID-19 relevant contents. & Tasks related to COVID-19. \\
           \midrule
       RuDR-BERT  \citep{Tutubalina2020} & Multilingual BERT adapted to pharmacology in Russian &  mBERT further pretrained on consumer reviews about pharmaceutical products & NER and multiclass classification. \\
            \bottomrule
 
    \caption{Representation models (i.e. embeddings for text and biological sequences) useful for pharmacology.}
    \label{tab:embeddings}
\end{longtable} 

\subsection{Injecting Pharmacological Knowledge into Deep Neural Networks}
\label{sec:knowledgeInjection}
 
While large pretrained language models have significantly increased the performance of machine learning approaches for most NLP tasks, many shortcomings still make the approaches less robust as desired. Examples of weaknesses are processing of negation, uncertainty about factual knowledge, and lack of problem-specific knowledge \citep{rogers2020primer}.

The knowledge injection approaches attempt to address the shortcomings of large pre-trained models by utilizing external knowledge resources in various forms, such as knowledge graphs (KGs, see Section \ref{sec:knowledgegraphs}) and other types of knowledge bases. This can reduce the need for ever-larger language models while improving their interpretability. In general, knowledge injection approaches differ in time of injection (during a pretraining phase, as an intermediate task, or in a downstream task), type of injected knowledge (facts, linguistic knowledge, commonsense reasoning, etc.), and type of evaluation (general language, domain-specific language, or probing).  

To improve pretrained language models for the biomedical domain, the existing approaches usually use the Unified Medical Language System (UMLS) knowledge base. UMLS is a medical terminology database with hundreds of biomedical vocabulary entries, including definitions of terms and relationships between them. The basic BERT model \citep{devlin2019bert} or any of the specific biomedical BERT models mentioned in Section \ref{sec:BERTs}, are used as a baseline where the knowledge is injected. 

Below, we present several approaches to knowledge injection in pharmacology. We divide them into the ones that modify existing pretraining tasks with general improvements in mind and those that focus on better concept representation for a specific task. We summarize the presented models in Table \ref{tab:KE_models}.

\subsubsection{Modification of Existing Pre-training Tasks for General Improvement}

This group of knowledge injection approaches focuses on developing new pre-training tasks or adding new modules to existing pretrained LMs. 

\citet{hao2020enhancing} improve biomedical LMs for medical downstream tasks by infusing the knowledge base information into the pretraining phase of the Clinical BERT. The authors used the MIMIC-III dataset and continued pre-training on the masked language modeling task and next sentence prediction. They also introduced the task of predicting whether a relationship exists between two concepts in the UMLS knowledge base. Positive instances for this task are taken from the existing relations in ULMS, while negative ones are created through negative sampling as relations in ULMS are very sparse. The final loss function used in training is a combination of all three tasks. The resulting knowledge-enhanced Clinical BERT was evaluated on two named entity recognition datasets and one natural language inference dataset, and the results showed an improvement over the baseline biomedical models BioBERT and Clinical BERT.

UmlsBERT \citep{michalopoulos2021umlsbert} also integrates external knowledge resources to improve biomedical language models. The authors updated the masked language modeling in the pre-training step with the associations between the words specified in the UMLS. Firstly, at the input level, medical terms are enhanced by their semantic types (UMLS contains 44 unique semantic types). For example, the model receives information that `lungs' are `body part', `organ' etc. This represents an additional input layer that must be trained. Words without semantic type are represented by a zero-filled vector. Secondly, the masked language modeling (MLM) task is modified: instead of predicting one missing token, the model predicts all words associated with the same concept unique identifier (CUI). For instance, where the standard MLM task predicts only `lung', the modified one predicts `lungs' and `pulmonary' as well. UmlsBERT achieves the best results in four out of the five tasks (one NLI and four NER tasks). The ablation study checking if semantic type information improves the performance shows that the model performs significantly worse on all tasks without it. 

\citet{meng2021mixture} improve biomedical BERTs by partitioning a very large KG into smaller subgraphs and infusing this knowledge into various BERT models using adapters. Adapters \citep{houlsby2019parameter} are BERT additions that add only a few new trainable parameters while the original weights remain fixed. This reduces the inefficiency of fine-tuning large models for each task and allows a high degree of parameter sharing.  \citet{meng2021mixture} construct two KGs from the UMLS knowledge graph. The METIS algorithm \citep{karypis1998fast} partition the knowledge graph into $n$ subgraphs. Following that, they train an adapter module for each sub-graph to predict the tail entity of a triplet from the sub-graph. Finally, they use AdapterFusion mixture layers \citep{pfeiffer2021adapterfusion} to combine the knowledge from adapter modules. They experimentally determined that 20 sub-graphs and PubMedBERT yielded the best results. Their approach improves performance on QA, NLI, and document classification tasks in the biomedical domain.

\subsubsection{Improved Concept Representation for Specific Tasks}

This group of knowledge injection approaches focuses on improving concept representations for specific tasks. 

The same medical concepts can be represented by a variety of nonstandard names, misspellings, and abbreviations. Term normalization is a task that addresses this problem. CODER \citep{yuan2022coder} proposes dual contrastive learning simultaneously on both terms and relation triplets from the UMLS KG. The approach is motivated by examples such as that it is better to have ``rheumatoid arthritis'' closer to ``osteoarthritis'' than ``rheumatoid pleuritis'' because both are subtypes of arthritis. Relations between terms express that and thus provide useful information during the training. CODER maximizes similarities between positive term-term pairs and term-relation-term pairs from the KG. They evaluate their approach on datasets in different languages consisting of term normalization, relation classification, and conceptual similarity tasks. Their approach significantly outperforms existing medical embeddings in zero-shot term normalization.   

\citet{liu2021self} address the problem of entity linking, specifically, the heterogeneous naming of medical concepts. The authors pre-train a transformer-based language model on the UMLS biomedical KG. They propose a metric learning framework that learns to cluster synonyms of the same concept. The goal of a self-alignment pre-training step is to learn such concept embeddings that maximize the similarity between two concepts based on the cosine similarity measure. The learning setup consists of triplets in the form $(x_a, x_p, x_n)$, where $x_p$ is a positive match for $x_a$ and $x_n$ is its negative match. This approach first samples hard triplets (triplets that contain negative pairs closer in space than positive pairs with basic BERT embedding by some margin). It learns to push negative pairs away from each other and positive pairs together by considering the multi-similarity loss function. The resulting SAPBERT improves the accuracy across six medical entity linking tasks (up to 20\%) compared to the domain-specific BERT models and achieves state-of-the-art results.

\begin{longtable}[l]{p{3.0cm}|p{3.0cm}|p{3.4cm}|p{5.0cm}}
        \toprule
        \textbf{Name} & \textbf{External knowledge} & \textbf{Pretrained model} & \textbf{Evaluation tasks}\\
        \toprule
        \citet{hao2020enhancing}        & MIMIC-III, UMLS                            & ALBERT                                             & NER, NLI                                      \\
        \midrule
        UmlsBERT \citep{michalopoulos2021umlsbert} & MIMIC-III, UMLS                            & Bio\_ClinicalBERT                                   & NER, NLI                                      \\
        \midrule
        \citet{meng2021mixture}        & UMLS                                       & PubMedBERT                                         & document classification, NLI, QA              \\
        \midrule
        CODER \citep{yuan2022coder}                & UMLS                                       & PubMedBERT, mBERT                                  & term normalization                            \\
        \midrule
        SAPBERT \citep{liu2021self}                & UMLS                                       & PubMedBERT                                         & entity linking                                \\
        \midrule
        \citet{mao2020use}            & UMLS                                       & BioWordVec                                         & semantic relatedness, WSD \\
        \bottomrule
    \caption{Summary of knowledge enhanced models. All methods are evaluated on more than one pretrained model. Here we report the one that achieved the best results. For CODER, we report the best monolingual in multilingual versions of the models.}
    \label{tab:KE_models}
\end{longtable} 

\citet{mao2020use} tackle the problem of measuring semantic relatedness between biomedical concepts (UMLS concepts). Semantic similarity expresses the relatedness of two concepts in their meaning and is an important tool for automatic spelling correction, information retrieval, and word sense disambiguation. Authors use pre-trained word embedding models (e.g., BioWordVec \citep{zhang2019biowordvec}, variations of BERT, etc.) to generate concept sentence embeddings from UMLS, and various graph embedding models (e.g., GCNs \citep{kipf2016semi}, TransE \citep{bordes2013translating} and its variants). In addition to that, they combined both concept sentence embeddings and graph embeddings by concatenation. The similarity score between two embeddings was computed using the cosine similarity measure. The combined word and graph embeddings produced the best results on three semantic relatedness datasets and a one-word sense disambiguation dataset.

\subsection{Explainable NLP in Pharmacology}
\label{sec:explanation}

Deep learning models commonly surpass standard machine learning models in terms of predictive performance. However, their decision-making process is typically opaque, meaning that it is difficult to explain why the model made a certain prediction. Understanding models' inner workings are helpful for debugging errors, possibly improving their performance, and  gaining scientific insights into the modeled process, e.g., why two drugs interact in the drug-drug interaction identification. Additionally, as pharmacology is concerned with drugs affecting humans, it is essential that predictions are safe and verifiable.

Depending on the time when an explanation is created, there are two types of explanation methods: intrinsic and post-hoc \citep{nlp-explanations-survey}. Intrinsic methods use a model’s architecture or its components to construct an explanation. A simple example is a linear regression model using binary bag-of-words features. The learned weights associated with the input words represent an explanation of the prediction for the given input (positive weights indicate the positive impact of words on the decision, and negative weights indicate negative impact). Another commonly used intrinsic method used for large pretrained transformer models is the inspection of attention weights, which intuitively represent the parts of the input the model focuses on. Attention, being the key component of the currently dominant transformer-based models, is easy to compute. However, multiple attention heads may be difficult to comprehend, and the alignment between attention explanations and the underlying model behavior (i.e. actual explanations) is questionable \citep{jain-wallace-2019-attention, wiegreffe-pinter-2019-attention}.

Post-hoc explanation methods construct an explanation after a model is trained. While intrinsic methods are based on the design of a specific model, post-hoc methods are typically model-agnostic. An example of such methods are perturbation-based explanation methods such as Local Interpretable Model-agnostic Explanations (LIME) \citep{ribeiro2016-lime}, SHapley Additive exPlanations (SHAP) \citep{lundberg2017-shap}, and Interactions-based Method for Explanation (IME) \citep{ime}. They work by repeatedly modifying (perturbing) the input, observing the changes in the output, and modeling their associations using a surrogate model. Post-hoc methods are convenient due to their flexibility in the choice of used model architectures. However, the faithfulness of the produced explanations may be poor \citep{slack2020-unfaithful, frye2021shapley} as they explain the model from an external perspective. 

Both intrinsic and post-hoc methods have been successfully applied to general \citep{lai-2019-text-explanations} and topic-specific language tasks in bio-medicine \citep{moradi-2021-biotext-explanations}. Below, we describe several cases of using explanation methods in  pharmacology. We provide an overview of the methods in Table \ref{tab:explainable-nlp-pharmacology}.

\citet{jha2018-interpretable-embeddings} make pre-trained word embeddings more interpretable by learning a transformation to a more interpretable embedding space with the retained performance. The interpretable word embeddings correspond to categorical embeddings, trained separately using expert-provided definitions and additional knowledge from a biomedical knowledge graph.

\citet{wawrzinek2020-dda-analogies} introduce an entity embedding-based explanation method for drug-disease association (DDA) prediction. They construct explanations following the drug-centric and the disease-centric notion of similarity:
\begin{itemize}
    \item drug-centric: ``if two drugs are chemically similar, they likely have a similar relationship with the target disease'';
    \item disease-centric: ``a drug has the same relation for similar diseases''.
\end{itemize}
To obtain the explanation, they embed the drug and the disease from a DDA pair and retrieve k intermediate entities (drugs or diseases) using a cosine similarity-based metric. An explanation instance is created based on the relationship between the drug, disease, and the intermediate entity in existing publications. For example, if the intermediate entity is a drug and the intermediate drug treats the input disease, the input drug is assumed to also treat the input disease with confidence proportional to the embedding similarity. The obtained explanations using $k$ intermediate entities are aggregated into the final DDA prediction, e.g., using a majority vote. 

\citet{huang2020caster} include an interpretable component in their drug-drug interaction (DDI) prediction system. The component projects the latent embedding of the input drug pair into a more interpretable subspace, whose basis consists of frequently occurring molecular substructures. The substructures are extracted from a database of drug representations by finding substrings with a high enough frequency. The projection into the subspace aims to capture the relevance of the molecular substructures towards the drug interaction prediction.

\citet{yazdani2022attentionsitedti} propose an attention-based drug-target interaction (DTI) prediction system, using the attention weights as an explanation. They demonstrate  the high predictive performance of their system on three benchmark datasets, while they demonstrate the interpretation capability of their model on a DTI prediction example via visualization.

\citet{bradshaw2019-molecule-chef} present a generator of product molecules from a set of common reactant molecules. It is composed of
\begin{itemize}
    \item an encoder-decoder model between a latent space and a list of reactant molecules, and 
    \item a reaction prediction model that transforms the reactants into a list of product molecules.
\end{itemize}
The second component introduces interpretability to the model as it provides some insight on \textit{how} the product molecules are constructed out of the reactants. However, the authors do not put an emphasis on evaluating the interpretability of their approach. 

\begin{longtable}[l]{p{3.6cm}|p{2.0cm}|p{4.2cm}|p{4.8cm}}
\toprule
\textbf{Reference} & \textbf{Explanation type} & \textbf{Short description} & \textbf{Downstream tasks} \\
\toprule
\citet{jha2018-interpretable-embeddings} & 
intrinsic & Interpretable word embedding transformation & semantic concept categorization \\
\midrule
\citet{wawrzinek2020-dda-analogies} & 
intrinsic & Embedding arithmetic (analogies) & drug-disease association prediction \\
\midrule
\citet{huang2020caster} & 
intrinsic & Interpretable subspace & drug-drug interaction prediction \\
\midrule
\citet{yazdani2022attentionsitedti} & 
intrinsic & Interpretable component (attention weights) & drug-target interaction prediction \\
\midrule
\citet{bradshaw2019-molecule-chef} & 
intrinsic & Interpretable component (reaction predictor) & molecule generation \\
\midrule
\citet{jiang2019-example-explanation} & 
post-hoc & Present representative examples & detection of potential adverse medication effects \\
\midrule
\citet{rodriguezperez2020-shap-qsar} &
post-hoc & Out-of-the-box method (SHAP) & structure-activity
relationship modeling \\
\midrule
\citet{pope2019-graph-explanations} & 
intrinsic & Adapt out-of-the-box methods (gradient-based saliency, CAM, EB) & identification of biological molecular properties \\
\bottomrule
\caption{An overview of the explanation methods used in NLP for pharmacology.}
\label{tab:explainable-nlp-pharmacology}
\end{longtable}

\citet{jiang2019-example-explanation} present an approach for detecting potential adverse medication effects from social media posts. The detection is posed as a word analogy task: given a known possible side effect of a drug, the task is to find similar pairs of drugs and corresponding side effects with a similar relation. The known possible side effects are taken from the SIDER database \citep{Kuhn2016-sider}, while the static word embeddings are trained on unlabeled tweets. The found potential side effects are subject to human examination along with relevant tweets expressing the effect.

\citet{rodriguezperez2020-shap-qsar} present a usability study of the SHAP explanation method for explaining complex compound activity prediction models. They find that SHAP produces consistent feature attributions across three complex models. Additionally, they demonstrate how the obtained attributions can be used to find potential biases in the models. 

\citet{pope2019-graph-explanations} present adaptations of three explanation methods for explaining graph convolutional neural networks: contrastive gradient-based saliency maps, class activation mapping, and excitation backpropagation (EB). They test the methods on molecular graph classification, where the task is to predict whether molecules possess certain properties, such as toxicity. The explanations are salient subgraphs, which can be interpreted as functional groups responsible for the molecular property (according to the model). By analyzing the explanations using automated metrics (fidelity, contrastivity, and sparsity), the authors conclude that the gradient-weighted class activation mapping is the most suitable out of the tested methods, although they emphasize the need for detailed studies of chemical validity of the explanations in future work. 

In summary, explanation methods have been adopted across a variety of pharmacology applications. We find that the authors typically use the explanation methods in one of two ways, either using the explanations as a safety mechanism for a semi-automatic use of the model predictions, or as a way to obtain plausible hypotheses that are then manually verified, for example using additional experiments. The proposed explanation methods for pharmacology commonly use a connection to an external knowledge source. We believe that the incorporation of external knowledge into explanation methods is a promising direction for further research as the prediction may not be intuitively explainable to  humans in terms of only input components. In addition, external human-curated knowledge may naturally be more intuitive to end-users.

\section{Common NLP Tasks and Applications}
\label{sec:NLPtasks}

Several NLP tasks are frequently tackled in the pharmacological context. Some of them are adapted from general NLP tasks (e.g., named entity recognition, relation extraction, and question answering). In contrast, others are specific to pharmacology (e.g., adverse drug reactions and literature-based drug discovery). 
We have mentioned some successful uses of contextual BERT models on these tasks in Section \ref{sec:BERTs}, but this mainly demonstrated the usability of these models. This section systematically analyzes the most important tasks in life sciences and pharmacology. As hundreds of works tackle these problems exclusively or among other problems, we review a sample of recent works. The overview is presented in Table \ref{tab:papers}.

\subsection{Named Entity Recognition for Pharmacology}
\label{sec:ner}

Named entity recognition (NER) – called entity identification, entity chunking, or entity extraction, is one of the most popular NLP techniques that classifies named entities in text into pre-defined categories such as person, time, location, organization, etc. In the biomedical context, the entities of interest can be cells, genes, gene sequences, proteins, biological processes and pathways, diseases, drugs, drug targets, compounds, adverse effects, metabolites, tissues, and organs \citep{perera2020named,bonner2021review}. NER is often used as the initial stage of analyses to provide semantic interpretations of unstructured text by identifying and categorizing concept references. Various concepts are detected with different degrees of difficulty. The critical issue in recognizing chemicals, for example, is the high variance in concept names and chemical formulas. In contrast, the main challenge in identifying gene functions is the high degree of uncertainty caused by species diversity.

In pharmacology domain, NER is often used as the first step of the relation extraction task (see Section \ref{sec:re}) \citep{kadir2013overview,gu2016chemical} or adverse drug reactions task (see Section \ref{sec:adr}) \citep{li2018extraction}. Many authors start with the MADE 1.0 challenge dataset, e.g.,  \citet{jagannatha2019overview} find the  medications and their attributes, \citet{chapman2019detecting} apply the conditional random field method for medication recognition,  \citet{yang2019madex} developed the MADEx model based on LSTM networks for the same purpose, and \citet{wunnava2019adverse} apply the Bi-LSTM model. 

\subsection{Relation Extraction for Pharmacology}
\label{sec:re}

The relation extraction task is part of information extraction (IE) and extracts semantic relationships from texts. The extracted relationships connect two or more entities of the same kind that fit into one of many semantic categories (for example, people, organizations, or places). Frequently, extracted relations are related to adverse drug reactions (ADR) and drug-drug interactions (DDI), relations between medications,  between their attributes such as dosage, route, frequency, and duration \citep{jagannatha2019overview}. The ability of NLP models to automatically detect adverse drug event (ADE) related terms in textual data helps avoid ADEs. This results in safer and better quality healthcare services, lower healthcare expenditures, more educated and engaged customers, and improved health outcomes.

In pharmacology, relation extraction typically processes scientific papers that provide novelties from the pharmacology. Classical approaches extracted semantic relationships with a pattern-based approach to find medical relations in pharmaceutical texts \citep{ben2011automatic,rosario2004classifying}. Deep learning approaches brought significant improvements \citep{li2018extraction,yang2019madex}. Lately used approaches apply pretrained language models, e.g.,  SemRep  \citep{kilicoglu2020broad}. The extracted information is sometimes used to construct graphs encoding drug-drug, and disease-drug relationships, representing the similarity between them \citep{zhou2020nedd}. Although most approaches are based on textual data, relations are also discovered through the analysis of EHR data \citep{chen2019robustly}. 

\subsection{Adverse Drug Reactions}
\label{sec:adr}

Adverse drug reaction (ADR) is defined as a considerably damaging or unpleasant reaction occurring from an intervention associated with the use of a pharmaceutical product. Adverse reactions frequently anticipate danger from future administration and demand avoidance, particular therapy, or dose regimen modification \citep{pirmohamed1998adverse}. ADRs have traditionally been divided into two categories. Type A responses are dose-dependent and predicted based on the drug's pharmacology (also known as enhanced reactions). In contrast, Type B responses, often known as weird reactions, are distinctive and unpredictable from the pharmacological point of view.

Implementation-wise, ADR extraction is similar to relation extraction, where ADRs connected to to various diseases and drugs are detected. Lately, large pretrained language models,such as BERT, are used in ADR extraction  \citep{breden2020detecting,li2020effective,hussain2021pharmacovigilance}. Again, texts are not the sole source of information, and EHRs are often used as additional information in ADR extraction \citep{li2018extraction,wunnava2019adverse,chapman2019detecting}.

\subsection{Literature Based Drug Discovery}
\label{sec:drugdiscovery}

LBD (literature-based discovery) is an automatic or semi-automatic method for discovering new information from the literature. The amount of scientific literature is steadily growing, driving researchers to become more specialized and making it challenging to track developments even in restricted fields \citep{henry2017literature}. If text is identified that overtly asserts the knowledge that "A is associated with B" and "B is associated with C" in the Swanson ABC co-occurrence model \citep{swanson1997interactive}, then the implicit knowledge of "A may be associated with C" is obtained. LBD is essential for biomedical NLP since it allows finding implicit information that can help to enhance biomedical research. A recent study presents the computational strategies utilized for LBD in the biomedical area \citep{gopalakrishnan2019survey}

LBD applies several NLP tasks to process the pharmacological and medical literature, with the purpose to detect new medical entities \citep{wang2020comprehensive,sang2018sematyp,dobreva2020improving}, extract relations \citep{preiss2015exploring,wang2017large} or reactions \citep{zhou2017optimizing}. Some approaches use scientific texts for protein engineering, and visualization \citep{biswas2021low}. Frequent information source is the PubMed engine together with the PubTator model \citep{wei2019pubtator} for automated annotation. The PharmKE tool \citep{jofche2021pharmke} labels pharmaceutical entities and the relationships between them. In new diseases, such as COVID-19, LBD technique have proved useful to extract relevant information \citep{pinto2020ace2,martinc2020covid}.  Another frequent task is  \textbf{drug repositioning} which helps to find another purpose for existing drugs, i.e., to use them in treating similar diseases \citep{xue2018review}. Alternatively, novel drug indications can be discovered by analyzing the medical history, as exemplified in the PREDICT model \citep{gottlieb2011predict}. 

\begin{longtable}[l]{p{5cm}|p{5cm}|p{5cm}}
    
        \toprule
        \textbf{Task} & \textbf{Description}&\textbf{Referenced papers}\\
        \toprule
        \textbf{Named Entity Recognition} & Identifying pharmaceutical entities in textual data & \citep{jagannatha2019overview} \citep{chapman2019detecting} \citep{wunnava2019adverse} \citep{gu2016chemical} \citep{kadir2013overview} \citep{li2018extraction} \citep{yang2019madex}\\
      
        \midrule
       \textbf{Relation Extraction} & Finding relation between drugs and diseases from scientific text resources &  \citep{ben2011automatic} \citep{li2018extraction} \citep{chen2019robustly} \citep{kilicoglu2020broad}  \citep{yang2019madex} \citep{rosario2004classifying}  \citep{zhou2020nedd}\\
        \midrule
      \textbf{Adverse Drug Reactions} &  Anticipate danger from future administration and
        demand avoidance, particular therapy, or dose regimen modification& \citep{breden2020detecting} \citep{li2020effective} \citep{hussain2021pharmacovigilance} \citep{li2018extraction} \citep{wunnava2019adverse} \citep{chapman2019detecting}\\
        \midrule
        \textbf{Literature Based Drug Discovery} &  Discovering new pharmacological information from
existing literature.& \citep{zhou2017optimizing} \citep{biswas2021low} \citep{wei2019pubtator} \citep{wang2020comprehensive} \citep{pinto2020ace2}\citep{martinc2020covid}\citep{sang2018sematyp} \citep{jofche2021pharmke} \citep{preiss2015exploring} \citep{wang2017large} \citep{dobreva2020improving} \citep{xue2018review} \citep{gottlieb2011predict}\\
        \midrule
        \textbf{Question Answering}&Answers given question with the most relevant response&\citep{su2020caire} \citep{lee2020answering} \citep{farrar2002arizona} \citep{veisi2020persian} \citep{marginean2014gfmed}\\
        \bottomrule
    \caption{Overview of tasks related to the pharmacology.}
    \label{tab:papers}
\end{longtable} 

\subsection{Question Answering}
\label{sec:QA}

Question answering (QA) is an NLP task that takes a question as input and returns an answer in the form of a ranked list of relevant replies, or a summary answer snippet \citep{coleman2005introducing}. In a classical (pre-neural) approach, QA incorporates three tasks: information retrieval, retrieving relevant documents or passages for a particular query, and text summarization that summarizes the reply from relevant passages. A related information retrieval task is called "Learning by Doing" and searches the knowledge base for entities most related to the ones mentioned in the question. This task is divided into ranking the texts found in the database and finding the correct answer among the recovered paragraphs.

QA can summarize the pharmacological literature, e.g., for new diseases like COVID-19 \citep{su2020caire}. The data are mainly from PubMed articles, and in the case of COVID-19, also news about this disease \citep{lee2020answering}. To answer pharmaceutical questions, the QA task can be applied in many languages, even in low-resource languages such as Persian \citep{veisi2020persian}. Another source of information can be linked data as used in the GFMed model \citep{marginean2014gfmed}.

\section{Data Resources}
\label{sec:data}

As the application of open science and open data principles is rising \citep{burgelman2019open}, the number of publicly available datasets is steadily growing. This makes finding and discovering appropriate datasets increasingly challenging. There are two strategies to find a dataset suitable for a given task. First, a bottom-up approach starts by searching available datasets and evaluating their utility for the given problem. Second, a top-down approach first finds relevant papers for the tackled topic and then explores the available datasets used in the papers. 

We first present an overview of specialized search engines for discovering and finding datasets in \ref{sec:searchEngines}. Then we give an overview of the most important datasets utilized in published papers related to NLP in pharmacology. The covered datasets are organized into five groups: patient data, drug usage data, drug structure data, question answering datasets, and general pharmacological data. In Section \ref{sub:patient_data}, we present datasets containing patients' history and medical notes. The datasets in Section \ref{sub:drug_usage_data} contain drug characteristics according to the prescriptions to patients, while in Section \ref{sub:drug_structure_data}, we cover datasets with information about drugs' chemical composition. Datasets supporting question answering systems in pharmacology are described in \ref{sub:qa_datasets}. Section \ref{sub:other_datasets} describes general resources useful for successful NLP in pharmacology. 

We include public and closed (private/commercial) data in the survey. The summary of datasets is contained in Table \ref{tab:datasets}, where for each dataset, we include a list of references where the dataset was used, a short description, the size of the dataset, and its typical usage.

\subsection{Finding and Discovering Datasets}
\label{sec:searchEngines}

As the number of datasets rapidly grows, it becomes essential to have effective tools for finding them. As a solution, there are several specialized search engines for discovering and finding datasets.

\textbf{Google's Dataset Search} (\url{https://datasetsearch.research.google.com/}) currently indexes more than 30 million publicly available datasets. Filters can limit the results based on licensing (free or premium), format (CSV, images, etc. ), and update time. Alternatively, a specialized cloud platform \textbf{data.world} (\url{https://data.world/}) hosts an enterprise data catalog with over 130,000 datasets and knowledge graphs. Another platform hosting public datasets is \textbf{Kaggle} (\url{https://www.kaggle.com/datasets}), which is primarily a machine learning competition platform, but it also includes a dataset search engine.

The NLP community usually publishes the source code and datasets in the \textbf{Github} (\url{https://github.com/}) repository so that this source control platform can be used for dataset discovery. A specialized platform indexing the code and data related to research papers is \textbf{Papers with Code} (\url{https://paperswithcode.com/datasets}). This platform offers research area-based organization of papers allowing for a convenient discovery and browsing of papers and datasets.
One of the most popular development platforms for NLP, the \textbf{Huggingface}, offers a good dataset search engine organized by NLP task, category, language, size, and license (\url{https://huggingface.co/datasets}). 

A specialized search engine for linked data is the \textbf{Linked Open Data (LOD) Cloud} (\url{https://lod.openlinksw.com/}) that allows for text-based search and entity lookup. LOD Cloud is a distributed web of interconnected datasets (over \num{1500} datasets) containing open data in a structured and semantically annotated format from multiple domains - life sciences, publications, government, media, etc. The background on the LOD Cloud is described in Section \ref{sec:knowledgegraphs}. 

\subsection{Patient Data}
\label{sub:patient_data}

Datasets with the information about patients typically contain patients' medical history or medical notes about them. The main application of these datasets is to find novel relations between drugs and diseases. Below, we briefly describe the most commonly used patient datasets.

\textbf{MIMIC-III: Medical Information Mart for Intensive Care} \citep{johnson2016mimic} (\url{https://mimic.mit.edu/}) is a dataset that contains data on patients hospitalized in large tertiary care hospitals critical care units. It contains information on vital signs, medicines, laboratory measurements, care providers' observations and notes, fluid balance, procedure codes, diagnostic codes, imaging reports, hospital length of stay, survival statistics, etc. This dataset contains data on over \num{40000} patients.

\textbf{MADE1.0 Database} \citep{data-competition-made1.0} (\url{https://bio-nlp.org/index.php/announcements}) is a Electronic Health Record (EHR) database that is a part of the MADE1.0 competition. The structured dataset contains information on taken drugs, experienced ADEs (Adverse Drug Events), and indications and symptoms of patients. The competition addressed three tasks: NER, relation identification, and a joint NER-RI task. The dataset contains \num{1089} patient notes with detailed named entity and relation annotations. 

\textbf{n2c2 NLP Research Database} \citep{henry20202018} (\url{https://portal.dbmi.hms.harvard.edu/projects/n2c2-nlp/}) is database used for Track 2 of the 2018 National NLP Clinical Challenges shared task. The data is extracted from the MIMIC-III (Medical Information Mart for Intensive Care-III) clinical care database. The records were chosen using a query that looked for ADEs in the description of records' ICD (International Classification of Diseases) code. The retrieved records were manually inspected to ensure that at least one ADE was present and adequately annotated. The dataset contains 505 discharge summaries in textual format.

\textbf{MarketScan} \citep{adamson2008health} (\url{https://www.ibm.com/products/marketscan-research-databases}) dataset is a collection of administrative claims databases that includes information on in-patient and out-patient claims, out-patient prescription claims, clinical usage records, and healthcare costs in US. The three main databases each contain a convenience sample for one of the following patient populations: (1) employees with contributing employers' health insurance, (2) Medicare beneficiaries with employer-paid supplemental insurance, and (3) Medicaid recipients in one of eleven participating states. The data is not in textual format but can be used with NLP applications. The database contains data on approximately 43,6 million persons. 

\subsection{Drug Usage Data}
\label{sub:drug_usage_data}

Datasets described in this section provide information on drugs' usage, usage instructions, effects,  pharmaceutical properties, and composition. 

\textbf{DailyMed Database} \citep{national2014dailymed} (\url{https://dailymed.nlm.nih.gov/dailymed/index.cfm}) is a web database provided by the National Library of Medicine (NLM) in US. The US Food and Drug Administration (FDA) updates the material daily. The DailyMed contains prescription and nonprescription medications for human and animal usage, medical gases, gadgets, cosmetics, nutritional supplements, and medical foods. The labeled drugs describe the composition, form, packaging, and other properties of drug products according to the HL7 Reference Information Model (RIM). These details are given in the descriptive text format. The database contains \num{142981} labels.

\textbf{DrugBank Database} \citep{wishart2018drugbank} (\url{https://go.drugbank.com/}) is one of the biggest drug databases. Besides drugs, it contains drug paths which show how the drug travels in the human body and allows search for indications and drug targets. For an individual drug, the database contains all the brand names, background information in the text form, its type, structure, weight, formula, other names it is called by, what it is used for, what therapies it is used in, indications, doses, interactions, etc. All the details for each drug are available online and given in the descriptive text format.
The database contains descriptions of \num{14665} drug entries.

\subsection{Drug Structure Data}
\label{sub:drug_structure_data}

Datasets covered in this section contain drug characteristics regarding their chemical composition. Mainly, they are used for discovering new drugs or finding protein-protein interactions between drugs.

\textbf{ChEMBL Database} \citep{gaulton2012chembl} (\url{https://www.ebi.ac.uk/chembl/}) is an open-source database that contains binding, functional, and ADMET (Chemical absorption, distribution, metabolism, excretion, and toxicity) data for a wide range of drug-like bioactive chemicals. These data are regularly manually extracted from the published literature, then selected and standardized to enhance their quality and usability across a variety of chemical biology and drug-discovery research uses. The database includes 2.4 million bioassay measurements spanning \num{622824} chemicals, including \num{24000} natural products. The contents were produced by sifting through over \num{34000} papers published in twelve medicinal chemistry journals. The data from the journals containing details can also be used.

\textbf{UMLS: The Unified Medical Language Database} \citep{bodenreider2004unified} (\url{http://umlsks.nlm.nih.gov}) is a database of biomedical vocabularies. The NCBI (National Center for Biotechnology Information) taxonomy, Gene Ontology, MeSH (Medical Subject Headings), OMIM (Online Mendelian Inheritance in Man), and the Digital Anatomist Symbolic Knowledge Base are all included in the UMLS MetaThesaurus. The UMLS is not a textual database but is frequently used in NLP tasks, such as extracting concepts, relationships, or knowledge of pharmacological entities from texts. The UMLS has about 2 million names for over \num{900000} concepts from over 60 biomedical vocabularies and 12 million relationships between them. 

\textbf{PDB: The Protein Data Bank Database} \citep{bank1971protein} (\url{http://www.rcsb.org/pdb/}) is a global repository of structural data for biological macromolecules. To obtain the data, depositors used X-ray crystal structure determination, NMR (Nuclear magnetic resonance), cryo-electron microscopy, and theoretical modeling. The search queries also return the literature from which the data is extracted, e.g., the abstracts from medical articles that can be further used for NLP. The number of papers accessible in the textual format is not available, but the database contains \num{133920} Biological Macromolecular Structures, each accompanied by a related abstract.

\textbf{ChemProt Database} \citep{taboureau2010chemprot} (\url{https://biocreative.bioinformatics.udel.edu/news/corpora/chemprot-corpus-biocreative-vi/}) is a biology annotated database  based on several chemical-protein annotation resources, together with disease-associated protein-protein interactions (PPIs). ChemProt was utilized in the BioCreative VI text mining chemical-protein interactions shared task. The data contains PubMed abstracts in textual format together with annotated entities and interactions. The database has \num{1820} abstracts.

\subsection{Question Answering Data}
\label{sub:qa_datasets}

This section covers some datasets that can be used to build pharmacological question answering models.

\textbf{MQP Database} \citep{mccreery2020effective} (\url{https://github.com/curai/medical-question-pair-dataset}) comprises \num{3048} question-answer pairs that are categorized as similar or distinct by medical experts (i.e. not particular to COVID-19). Two doctors collaborated on the annotation and their agreement on \num{836} question pairings in the test set was above 85\%.

\textbf{COVID-Q Database} \citep{wei2020people} (\url{https://paperswithcode.com/dataset/covid-q}) is a collection of \num{1690} COVID-19-related questions divided into 15 general categories and 207 specific question classes. The dataset was annotated in three stages by many curators. First, two curators discussed and categorized the questions. Second, an external curator reviewed the work and, if necessary, proposed adjustments to the categories. Third, questions from more than four different question classes were sampled and allocated to three different AMT (Amazon Mechanical Turk) workers. The validation was based on the majority vote.

\textbf{CovidQA Database} \citep{zhao20202019} (\url{https://aclanthology.org/2020.nlpcovid19-acl.18/}) is made up of 124 question–article–answer triplets taken from 85 different articles in CORD-19 Kaggle challenge and covers 27 different categories. Five curators created annotations by synthesizing questions from the challenge organizers' categories, then manually discovered relevant articles and replies.

\subsection{General Pharmacological Data}
\label{sub:other_datasets}

In this section, we describe five resources that are general and useful for many tasks. 

\textbf{Wikipedia} \citep{wikipedia2004wikipedia} (\url{https://en.wikipedia.org/}) is a well known encyclopedia and  web-based collaborative database consisting of over 15 billion articles. Wikipedia contains articles from different scientific fields written in many languages.

\textbf{PubMed} \citep{canese2013pubmed} (\url{https://pubmed.ncbi.nlm.nih.gov/}) is a free web engine for primarily MEDLINE, bibliographic database encompassing medicine, nursing, dentistry, veterinary medicine, the health-care system, and preclinical sciences like molecular biology. More than \num{4600} biomedical journals are indexed in MEDLINE, together with bibliographic citations and author abstracts. PubMed indexes more than 30 million articles and abstracts.

\textbf{LitCovid Database} \citep{chen2021litcovid} (\url{https://www.ncbi.nlm.nih.gov/research/coronavirus/}) is a curated literature site for tracking up-to-date scientific knowledge regarding the COVID-19 disease. It is the most comprehensive resource on the topic with central access to more than \num{255935} relevant PubMed articles. The articles are updated daily and divided into categories based on research themes and geographical areas.

\textbf{CORD-19} (COVID-19 Open Research Database) \citep{wang2020cord} (\url{https://www.kaggle.com/datasets/allen-institute-for-ai/CORD-19-research-challenge}) contains metadata about papers related to COVID-19. The main sources are PubMed, World Health Organization, bioRxiv and medRxiv. This database contains over \num{52000} papers.

\textbf{DBpedia} \citep{auer2007dbpedia} (\url{https://www.dbpedia.org/}) is a structured open-source database with information extracted from Wikipedia articles. For drugs, it contains basic information on uses, contained chemicals, drug type, links to other languages, Wikipedia links, and other links used to extract information. The database contains more than \num{10000} drug type entries. 
 
\textbf{EMBASE} (Excerpta Medica dataBASE) (\url{http://www.Embase.com}) is a biological and pharmacological bibliographic database of published literature. It was created to assist information managers and pharmacovigilance in adhering to the regulatory requirements of a licensed medicine. Embase database, created in 1947, contains more than 32 million entries from more than 8,500 published journals.

\textbf{ClinicalTrials.gov} \citep{zarin2011clinicaltrials} (\url{https://www.clinicaltrials.gov/}) is a clinical trial registry and the biggest clinical trials database. It is managed by the National Institutes of Health and contains registrations for over 329,000 studies from 209 countries.

\begin{longtable}[l]{p{4.2cm}|p{4.6cm}|p{2.0cm}|p{3.6cm}}
        \toprule
        \textbf{Name} & \textbf{Description} & \textbf{Entries} & \textbf{Usage}\\
        \toprule
        \multicolumn{4}{c}{\textbf{Patient Data}}\\
        \midrule
        \endfirsthead
        MarketScan \citep{adamson2008health} & Collection of administrative claims & 43,600,000 & NER, ADE, Drug-drug interaction\\
        \multicolumn{4}{l}{URL: \url{https://www.ibm.com/products/marketscan-research-databases}}\\
        \midrule
        MIMIC-III \citep{johnson2016mimic} & Data on patients hospitalized & 40,000 &  Drug discovery, ADE, Drug-drug interaction\\
        \multicolumn{4}{l}{URL: \url{https://mimic.mit.edu/}}\\
        \midrule
        MADE 1.0 \citep{data-competition-made1.0} & A challenge dataset with 21 EHRs of cancer patients & 1,089 & NER, ADE\\
        \multicolumn{4}{l}{URL: \url{https://bio-nlp.org/index.php/announcements}}\\
        \midrule
         n2c2 \citep{henry20202018} & Unstructured notes from the Research Patient Data & 505 & ADE\\
         \multicolumn{4}{l}{URL: \url{https://portal.dbmi.hms.harvard.edu/projects/n2c2-nlp/}}\\
        \midrule
        \multicolumn{4}{c}{\textbf{Drug Usage Data}}\\
        \midrule
        DailyMed \citep{national2014dailymed}  & Drug label database & 142,981 & NER, Drug-drug interaction, ADE\\
        \multicolumn{4}{l}{URL: \url{https://dailymed.nlm.nih.gov/dailymed/index.cfm}}\\
        \midrule
        DrugBank \citep{wishart2018drugbank} & Database of drugs and drug products & 14,665 & ADE, pharmacovigilance, standardization, interactions \\
        \multicolumn{4}{l}{URL: \url{https://go.drugbank.com/}}\\
        \midrule
        \multicolumn{4}{c}{\textbf{Drug Structure Data}}\\
        \midrule
        ChEMBL \citep{gaulton2012chembl} & Binding, functional, and ADMET data & 2,400,000 & ADE, pharmacovigilance, standardization, interaction\\
        \multicolumn{4}{l}{URL: \url{https://www.ebi.ac.uk/chembl/}}\\
        \midrule
        UMLS \citep{bodenreider2004unified}  & Biomedical vocabularies & 2,000,000 & ADE \\
        \multicolumn{4}{l}{URL: \url{http://umlsks.nlm.nih.gov}}\\
        \midrule
        PDB \citep{bank1971protein} &  biological macromolecules & 133,920& ADE, pharmacovigilance, standardization, interaction\\
        \multicolumn{4}{l}{URL: \url{http://www.rcsb.org/pdb/}}\\
        \midrule
        ChemProt \citep{taboureau2010chemprot}  & Biological annotations &  1,820 & ADE\\
        \multicolumn{4}{l}{URL: \texttt{\href{https://biocreative.bioinformatics.udel.edu/news/corpora/chemprot-corpus-biocreative-vi/}{https://biocreative.bioinformatics.udel.edu/news/corpora/chemprot-corpus-}}}\\
        \multicolumn{4}{l}{\texttt{\href{https://biocreative.bioinformatics.udel.edu/news/corpora/chemprot-corpus-biocreative-vi/}{biocreative-vi/}}}\\
        \midrule
        \multicolumn{4}{c}{\textbf{Question Answering Data}}\\
        \midrule
        MQP \citep{mccreery2020effective}  & Collection of medical related pairs of questions and answers & 3,048 & QA\\
        \multicolumn{4}{l}{URL: \url{https://github.com/curai/medical-question-pair-dataset}}\\
        \midrule
        COVID-Q \citep{wei2020people}  &Collection of COVID-19-related questions divided into 15 general categories and 207 specific question classes &1,690 & QA\\
        \multicolumn{4}{l}{URL: \url{https://paperswithcode.com/dataset/covid-q}}\\
        \midrule
        CovidQA \citep{zhao20202019} &
        Collection of question–article–answer triplets taken from 85 different articles in CORD-19 & 124 & QA\\
        \multicolumn{4}{l}{URL: \url{https://aclanthology.org/2020.nlpcovid19-acl.18/}}\\
        \midrule
        \multicolumn{4}{c}{\textbf{General Pharmacological Data}}\\
        \midrule
        Wikipedia \citep{wikipedia2004wikipedia} & Online free encyclopedia & 15,000,000,000 & ADE, Drug-drug interaction, Drug discovery, NER\\
        \multicolumn{4}{l}{URL: \url{https://en.wikipedia.org/wiki/Main_Page}}\\
        \midrule
        PubMed \citep{canese2013pubmed} &Web engine for searching health articles & 30,000,000 & ADE, Drug-drug interaction, Drug discovery, NER\\
        \multicolumn{4}{l}{URL: \url{https://pubmed.ncbi.nlm.nih.gov/}}\\
        \midrule
        LitCovid \citep{chen2021litcovid} & Scientific PubMed articles related with COVID-19 & 255,935& ADE, Drug-drug interaction, Drug discovery, NER\\
        \multicolumn{4}{l}{URL: \url{https://www.ncbi.nlm.nih.gov/research/coronavirus/}}\\
        \midrule
        CORD-19 \citep{wang2020cord} & Scientific papers relevant to COVID-19 research & 52,000 & ADE, Drug-drug interaction, Drug discovery, NER\\
        \multicolumn{4}{l}{URL: \url{https://www.kaggle.com/datasets/allen-institute-for-ai/CORD-19-research-challenge}}\\
        \midrule
        DBpedia \citep{auer2007dbpedia} & Articles and structured data on e.g., drugs and diseases & 10,000 & ADE, pharmacovigilance, standardization, interactions \\
        \multicolumn{4}{l}{URL: \url{https://www.dbpedia.org/}}\\
        \midrule
        EMBASE & Biological and pharmacological bibliographic database & 32,000,000 & ADE, pharmacovigilance, standardization, interactions \\
        \multicolumn{4}{l}{URL: \url{http://www.embase.com}}\\
        \midrule
        ClinicalTrials.gov \citep{zarin2011clinicaltrials} & Clinical trials database & 329,000 & ADE, pharmacovigilance, standardization, interactions \\
        \multicolumn{4}{l}{URL: \url{https://www.clinicaltrials.gov/}}\\
        \bottomrule
    \caption{Different types of pharmacology-relevant datasets.}
    \label{tab:datasets}
\end{longtable}

\section{Knowledge Graphs}
\label{sec:knowledgegraphs}

The concepts of linked data and knowledge graphs introduced new standards for representing, storing, and retrieving data over the Web, both publicly and privately \citep{bizer2008linked,bizer2009linked,heath2011linked,wood2014linked,10.1145/3447772}. As a result of years of adoption of the linked data principles by various data publishers, the Linked Open Data (LOD) Cloud (\url{https://lod-cloud.net}) has been created and populated with \num{1541} interlinked datasets from the domains of geography, government, life sciences, linguistics, media, publications, social networking, user-generated, and cross-domain.

Knowledge graphs, the latest trend in the semantic web and linked data, enable the generation, consolidation, and contextual linking of structured data. The standards and technologies for knowledge graphs solve the problem of having separate `data silos' in traditional relational database systems, which have to be explicitly mapped to other isolated databases to take advantage of interconnected data \citep{jovanovik2017consolidating}.

\begin{figure}[!ht]
  \centering
  \includegraphics[width=\textwidth]{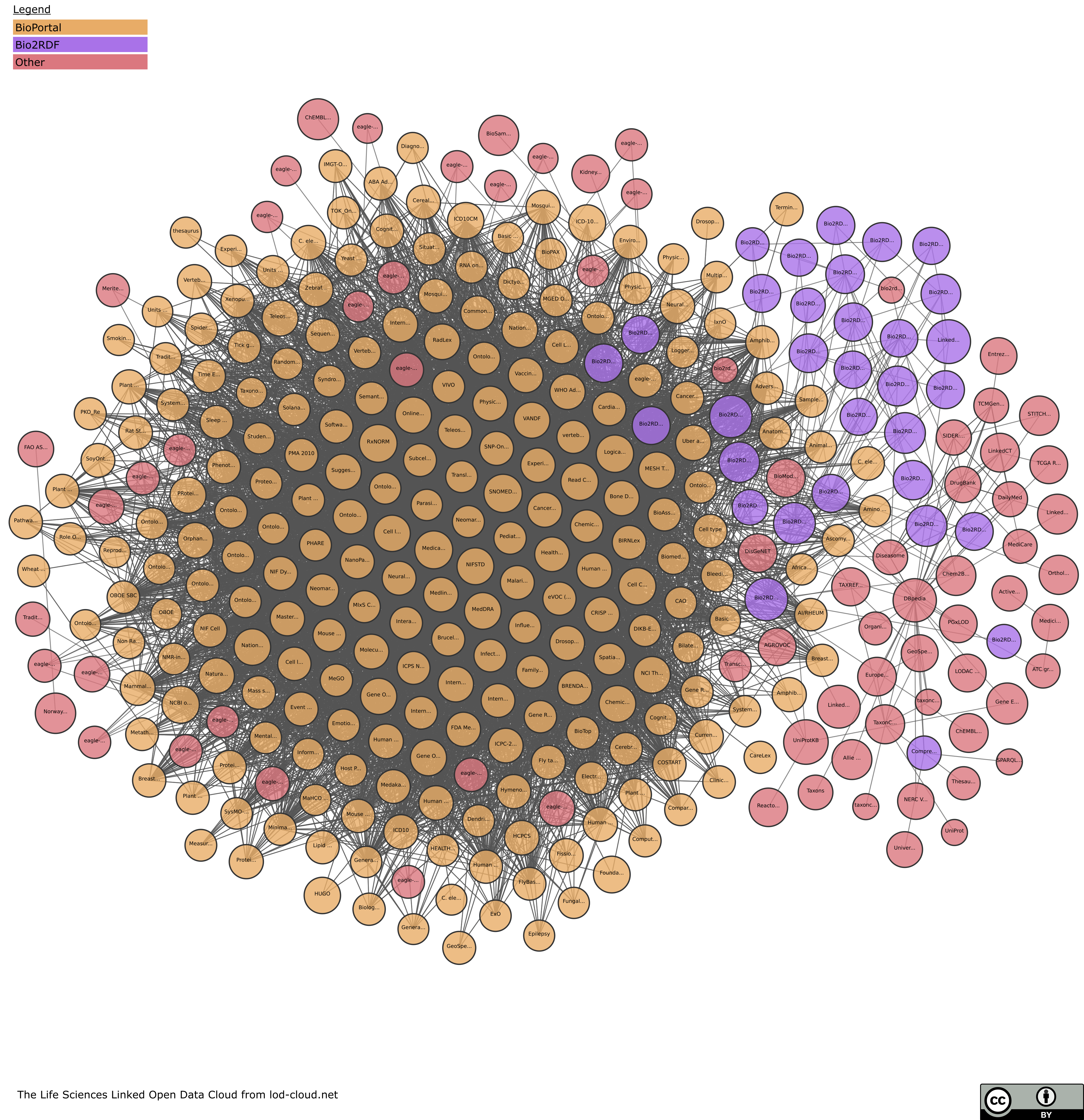}
  \caption{The Life Sciences sub-graph of the Linked Oped Data (LOD) Cloud, as of November 2022. Each node is a dataset. The weighted links denote the amount of RDF triples which link entities from the connected datasets.}
  \label{fig:lodcloud}
\end{figure}

The pharmaceutical industry is leading in using knowledge graph-based NLP techniques, especially in patient disease identification, clinical decision support systems, and pharmacovigilance \citep{Dumitriu2021}. The problem of identifying patients with specific diseases can be mitigated by knowledge graphs generated from structured and unstructured data from medical records, which capture explicit disease–symptom relationships \citep{chen2019robustly}. Recently, knowledge graphs improved the classification of rare disease patients \citep{8904588}. In the area of clinical decision support, the combination of NLP and knowledge graphs is employed in inferring drug-related knowledge which is not immediately observed in data, inferring cuisine-drug interactions based on knowledge graphs of drugs and recipes, improving user interaction with relevant medical data, etc. \citep{10.1145/2983323.2983819, jovanovik2015inferring, liu2018t, ruan2019qanalysis, xia2018mining, xia2022lingyi}. In pharmacovigilance, the struggles of NLP engines to understand complex language components (e.g., negation, doubt, historical medical statements, family medical history, etc.) from individual case study reports have been significantly mitigated with the use of knowledge graphs \citep{perera2013challenges}. Other examples include the use of knowledge graphs to improve NLP pipelines for detecting medication and adverse drug events from EHRs \citep{ngo2018knowledge}, as well as from Medline abstracts \citep{yeleswarapu2014pipeline}.

Section \ref{sec:biomedical-kgs} presents several knowledge graphs from the biomedical domain used in the mentioned application areas. Given the ongoing COVID-19 pandemic, we outline several recent COVID-19-related knowledge graphs in Section \ref{sec:CovidKG}. 

\subsection{Biomedical Knowledge Graphs}
\label{sec:biomedical-kgs}

Several projects worked on the transformation of pharmacology-related and healthcare data into linked data and knowledge graphs. Currently, 341 life science datasets are present in the LOD Cloud. These datasets contain healthcare data from various subdomains, such as drugs, diseases, genes, interactions, clinical trials, enzymes, etc. The most notable of them are presented below, and are outlined in Table \ref{tab:knowledgegraphs}.

\textbf{Bioportal} \citep{whetzel2011bioportal} (\url{https://bioportal.bioontology.org/}) project hosts ontologies covering drugs, diseases, genes, clinical procedures, etc. With over \num{980} biomedical ontologies, which define a total of over \num{13900000} classes, it represents the largest such repository in the life-science domain.

\textbf{Bio2RDF} \citep{callahan2013bio2rdf} (\url{https://bio2rdf.org}) is an open-source project which creates RDF datasets from various life science resources and databases and interconnects them into one network \citep{belleau2008bio2rdf,callahan2013bio2rdf,callahan2013ontology}. The latest release of Bio2RDF contains around 11 billion triples which are part of 35 datasets. These datasets contain various healthcare data: clinical trials (ClinicalTrials), drugs (DrugBank, LinkedSPL, NDC), diseases (Orphanet), bioactive compounds (ChEMBL), genes (GenAge, GenDR, GOA, HGNC, HomoloGene, MGD, NCBI Gene, OMIM, PharmGKB, SGD, WormBase), proteins (InterPro, iProClass, iRefIndex), gene-protein interactions (CTD), biomedical ontologies (BioPortal), side effects (SIDER), terminology (Resource Registry, MeSH, NCBI taxonomy), mathematical models of biological processes (BioModels), publications (PubMed), etc.

\textbf{Macedonian drug data}. Drug data from the Health Insurance Fund of North Macedonia has been transformed into a knowledge graph and linked to other LOD Cloud datasets \citep{jovanovik2013linked}. This knowledge graph was further extended with linked data about Macedonian medical institutions, and drug availability lists from pharmacies \citep{jovanovik2015linked}. 

\textbf{Cuisine - Drug interactions}. This project used two knowledge graphs for analysis of connections between drugs and their interactions with food, and recipes from different national cuisines, resulting in findings that uncovered the ingredients and cuisines most responsible for negative food-drug interactions in different parts of the world (\url{http://viz.linkeddata.finki.ukim.mk}) \citep{jovanovik2015inferring}.

\textbf{Global drug data}. In this research project, a pipeline-based platform was created to collect, clean, align, consolidate, and create a publicly available knowledge graph of drug products registered in various countries (\url{http://drugs.linkeddata.finki.ukim.mk}) \citep{jovanovik2017consolidating}. The source of the data is the official country drug registers. The generated RDF knowledge graph is publicly available through a web-based app (\url{http://godd.finki.ukim.mk}).

\subsection{COVID-19 Knowledge Graphs}
\label{sec:CovidKG}

COVID-19 pandemic turned the attention of many researchers to life sciences and healthcare domains. Below we list some recent COVID-19-related knowledge graphs.

\textbf{TypeDB Bio (Covid) knowledge graph} (\url{https://github.com/typedb-osi/typedb-bio}) contains data extracted from COVID-19 papers and from datasets on proteins, genes, disease-gene associations, coronavirus proteins, protein expression, biological pathways, and drugs. For instance, it allows querying for specific viruses giving associated human proteins related to the virus (e.g., a protein that helps in the replication of the virus). From here, it is possible to identify drugs that inhibit the detected proteins, meaning they can be prioritized in research as potential treatments for patients with the virus. To check the plausibility of this association and the implications, the graph can be used to identify relevant papers in the COVID-19 literature where this protein has been studied.

\begin{figure}[!ht]
  \centering
  \includegraphics[width=\textwidth]{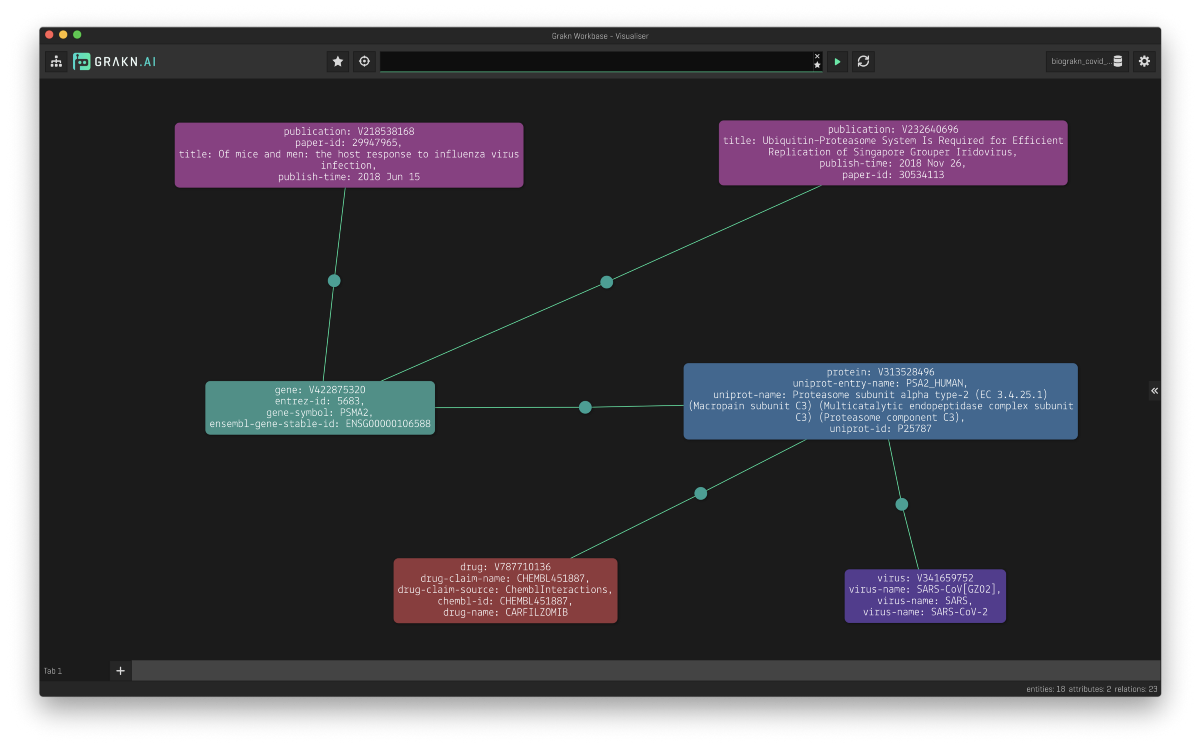}
  \caption{Example usage of the TypeDB Bio (Covid) knowledge graph.}
  \label{fig:typedbbiokg}
\end{figure}

\textbf{Covid-19-DS} \citep{pestryakova2022covidpubgraph} (\url{https://dice-research.org/COVID19DS}) is an RDF knowledge graph of scientific publications. The base of the graph is the CORD-19 dataset \citep{wang2020cord} that is regularly updated. The graph generation pipeline applies NER, entity linking, and link discovery to the CORD-19 data. The current version of the resulting graph contains over \num{69000000} RDF triples and is linked to 9 other datasets with over \num{1000000} links.

\textbf{KG-Covid-19} \citep{REESE2021100155} (\url{https://github.com/Knowledge-Graph-Hub/kg-covid-19/wiki})  is a framework that allows users to download and transform COVID-19 related datasets and generate a knowledge graph that can be used in machine learning. The project also provides access to pre-built knowledge graphs along with public querying. 

\begin{table}[!ht]
    \centering
    \begin{tabular}{l|r|r}
        \toprule
        \textbf{Name} & \textbf{Unique Entities} & \textbf{RDF Statements}\\
        \toprule
        Bio2RDF \citep{callahan2013bio2rdf} & 1,107,871,027 & 11,895,348,562 \\
        \midrule
        HIFM \citep{jovanovik2013linked, jovanovik2015linked} & 3,000 & 21,233 \\
        \midrule
        LinkedDrugs \citep{jovanovik2017consolidating} & 248,746 & 99,235,032 \\
        \midrule
        Covid-19-DS & 262,954 & 69,434,763 \\
        \midrule
        KG-Covid-19 \citep{REESE2021100155} & 574,778 & 24,145,556 \\
        \bottomrule
    \end{tabular}
    \caption{Covered knowledge graphs from the biomedical domain and their characteristics.}
    \label{tab:knowledgegraphs}
\end{table}

\section{Tools and Libraries}
\label{sec:tools}

This section focuses on the technical part of NLP applications in pharmacology. In Section \ref{sec:MLlibs}, we cover software libraries and tools that help to build machine learning models for the tasks mentioned in Sections \ref{sec:NLPmethodologies} and \ref{sec:NLPtasks}. For each library, we also mention its recorded use in pharmacology. In Section \ref{sec:GeneralLibs}, we present general text processing libraries. Most covered libraries and tools are accessible as Python packages. Table \ref{tab:libraries} gives an overview.

\subsection{Machine Learning Libraries}
\label{sec:MLlibs}

\textbf{Natural Language Toolkit (NLTK)} \citep{bird2009natural} (\url{https://www.nltk.org/}) is one of the most powerful and popular NLP libraries. NLTK is a suite of open-source Python modules, data sets, and tutorials on language processing. The toolkit consists of baseline text processing such as sentence splitting, tokenization, and part of speech (POS) tagging. These tools may help in NER, to identify known medications, detect ADEs \citep{chapman2019detecting}, or in evaluation of entity indicators for relation extraction \citep{qin2021entity}.

\textbf{MetaMap Transfer (MMTx)} \citep{aronson2001effective} (\url{https://github.com/theislab/MetaMap}) is an extensively used, Java-based, NER tool that maps biomedical free-form text to UMLS Metathesaurus concepts. In the process of creating the first drug-drug interaction (DDI) corpus that, besides drugs, contains pharmacokinetic  DDIs and pharmacodynamic DDIs, the UMLS MetaMap Transfer tool pre-annotates the documents with pharmacological substance entities, i.e., it is used to parse the documents to automatically recognize drug types \citep{herrero2013ddi}. MetaMap's intrinsic function - identification of medical concepts - was used for extracting drug indication information from structured product labels \citep{fung2013extracting}.
    
\textbf{CRFsuite} \citep{CRFsuite} (\url{http://www.chokkan.org/software/crfsuite/}) implements the Conditional Random Fields machine learning algorithm for labeling sequential data. It is used for NER in the MADEx system for detecting medications and ADEs and their relations from clinical notes \citep{yang2019madex}. 
    
\textbf{Library for Support Vector Machines (LibSVM)} \citep{CC01a} (\url{https://www.csie.ntu.edu.tw/~cjlin/libsvm/}) is an open-source package that implements the Sequential minimal optimization (SMO) algorithm for kernelized support vector machines (SVMs), supporting both classification and regression. The library was used to classify relation types in the MADEx system \citep{yang2019madex}.

\textbf{Stanford CoreNLP toolkit} \citep{manning2014stanford} (\url{https://stanfordnlp.github.io/CoreNLP/}) was initially developed for English, but now supports German, French, Arabic, Chinese and Spanish. The Stanford CoreNLP toolkit is a pipeline of NLP Java tools for linguistic annotations, such as tokenization, sentence splitting, part-of-speech tagging, morphological analysis, NER, syntactic parsing, and coreference resolution. In pharmacology, CoreNLP was applied in a joint model for entity and relation extraction from biomedical text, providing POS tagging and dependency parsing \citep{li2017neural}.
    
\textbf{BRAT annotation tool} \citep{stenetorp2012brat} (\url{https://brat.nlplab.org/introduction.html}) is an online environment for annotating structured text, i.e. notes in a predefined form. The tool was used to create a corpus from Twitter messages and PubMed sentences to understand drug reports better \citep{alvaro2017twimed}.
    
\textbf{SpaCy library} \citep{spacy} (\url{https://spacy.io/}) is a free, open-source library for NLP. It contains ML models for NER, POS tagging, dependency parsing, sentence segmentation, text classification, entity linking, morphological analysis, etc. The library is employed for entity recognition for Pharmaceutical Organizations and Drugs in PharmKE - a text analysis platform focused on the pharmaceutical domain \citep{jofche2021pharmke}.
    
\textbf{DOMEO Annotation Toolkit}  \citep{ciccarese2012domeo} (\url{https://github.com/domeo/domeo}) (also called SWAN Annotation Tool) is a web application enabling users to manually, semi-automatically, or automatically create ontology-based annotation metadata. DOMEO (Document Metadata Exchange Organizer) can be customized with additional plugins, e.g., for  annotation of PDDI mentions in structured product labels \citep{hochheiser2016using} (\url{https://github.com/rkboyce/DomeoClient}).

\textbf{Transformers - Hugging Face} \citep{wolf-etal-2020-transformers} (\url{https://huggingface.co/}) package contains many state-of-the-art NLP models, such as BioBERT \citep{Lee2019BioBERT}, RoBERTa \citep{liu2019roberta}, CharacterBERT \citep{el2020characterbert}, etc. The package offers also tokenizers for several languages and tasks, as well as some popular datasets for NLP tasks such as NER, NLI, QA, etc.
    
\textbf{MedCat Tool} \citep{kraljevic2019medcat} (\url{https://github.com/CogStack/MedCAT}) (Medical Concept Annotation Tool)  is an open-source tool that uses  unsupervised methods for NER and NEL in the biomedical field. The tools were validated with the MIMIC-III program and MedMentions (biomedical papers annotated with mentions from critical care databases). \citet{dobreva2022dd} highlighted drug entities with the help of this tool in the process of extracting drug-disease relations and drug effectiveness.  

\textbf{AllenNLP} \citep{gardner2018allennlp} (\url{https://allenai.org/allennlp}) is an open-source research library, built on PyTorch, for developing deep learning models for a wide variety of linguistic tasks. The PharmKe \citep{jofche2021pharmke} model uses AllenNLP for NER of drugs and pharmaceutical organizations that appear in texts.
    
\textbf{Flair} \citep{akbik2019flair} (\url{https://github.com/flairNLP/flair}) is a simple yet powerful framework for NLP, such as NER, POS tagging, and text classification. The framework supports training new models and is used in many research projects and industrial applications, e.g., \citet{sun2021deep} use FLAIR to find sub-word embeddings.

\textbf{Gensim} \citep{rehurek_lrec} (\url{https://radimrehurek.com/gensim/}) is a Python library for topic modeling -  extraction of unknown topics from a large volume of text (feeds from social media, customer reviews, user feedback, e-mails of complaints, etc.), document indexing, and similarity retrieval from large corpora. The library can handle large text files without having to load the entire file into memory, has efficient multicore implementations of popular algorithms, is platform-independent, and supports distributed computing. \citet{dobreva2020improving} apply Gensim to NER.

\subsection{General NLP libraries}
\label{sec:GeneralLibs}
    
\textbf{JIEBA tool} \citep{sun2012jieba} (\url{https://github.com/fxsjy/jieba}) supports Chinese word segmentation based on word frequency statistics with several functions such as POS tagging, TF-IDF weightig and TextRank keyword extraction. It was used to generate POS tags of words \citep{qin2021entity}.

\textbf{TextBlob} \citep{loria2018textblob} (\url{https://textblob.readthedocs.io/en/dev/}) is a simple Python library, built on top of NLTK and Pattern, that supports complex analysis and operations on text data. The library supports noun phrase extraction, POS tagging, sentiment analysis, classification (Naive Bayes, Decision Tree), tokenization, word and phrase frequencies, parsing, n-grams, word inflection (pluralization and singularization) and lemmatization, spelling correction, etc.

\textbf{Polyglot} \citep{nystrom2003polyglot} (\url{https://github.com/aboSamoor/polyglot}) is a NLP pipeline that supports multilingual applications and offers a wide range of analyses. It features tokenization (165 languages), language detection (196 languages), NER (40 languages), POS tagging (16 languages), sentiment analysis (136 languages), word embeddings (137 languages), morphological analysis (135 languages), and transliteration (69 languages).
    
\textbf{Quepy} \citep{andrawos2012quepy} (\url{https://github.com/machinalis/quepy}) is a Python framework to transform natural language questions to queries in a database query language.
  
In Table \ref{tab:libraries}, we overview the mentioned libraries, together with references from the papers where they are used.

\newcommand{\specialcell}[2][c]{%
  \begin{tabular}[#1]{@{}c@{}}#2\end{tabular}}

\begin{longtable}[l]{p{3.4cm}|p{5.6cm}|p{6.0cm}}
        \toprule
        \textbf{Name} & \textbf{Usage} & \textbf{Referenced Papers} \\
        \toprule
        Natural Language Toolkit (NLTK)\citep{bird2009natural} & Tokenization, Lemmatization, POS tagging, NER, Word similarity & \citep{segura2017simplifying} \citep{khadhraoui2022survey} \citep{jagannatha2019overview} \citep{liu2019pattern} \citep{aldahdooh2021r} \citep{chen2020extracting} \citep{li2020survey} \citep{chapman2019detecting} \citep{bird2009natural} \citep{turina2021thermoscan} \citep{sivasankari2017medical} \citep{prabadevi2019heart} \citep{mahatpure2019electronic} \citep{rabhi2019deep}\citep{ren2021variability} \citep{romasanta2020innovation} \citep{sjogren2020multivariate} \citep{raghupathi2018legal}\\
        \multicolumn{3}{l}{URL: \url{https://www.nltk.org/}}\\
        \midrule
        MetaMap Transfer tool (MMTx)\citep{aronson2001effective} & \specialcell{NER, DD Interaction}& \citep{schriml2012disease} \citep{aronson2001effective} \citep{ben2011automatic} \citep{fung2013extracting} \citep{gottlieb2011predict} \citep{sang2018sematyp} \citep{preiss2015exploring} \citep{kilicoglu2020broad} \citep{yang2011mining} \citep{jagannatha2019overview} \citep{yang2019madex} \citep{perera2020named} \citep{kamp2013application} \citep{mattes2013prediction} \citep{chiaramello2016use} \citep{jiang2013mining}\\
        \multicolumn{3}{l}{URL: \url{https://github.com/theislab/MetaMap}}\\
        \midrule
        CRFsuite library\citep{CRFsuite}&NER, Drug Discovery, ADE&\citep{pyysalo2013distributional} \citep{chapman2019detecting} \citep{yang2019madex} \citep{Habibi2017deep} \citep{bamburova2019structured} \citep{hakala2019biomedical} \citep{soysal2018clamp} \citep{ngo2018knowledge} \citep{liu2015effects}\\
        \multicolumn{3}{l}{URL: \url{http://www.chokkan.org/software/crfsuite/}}\\
        \midrule
        LibSVM\citep{CC01a} &\specialcell{Classification, Regression}& \citep{yang2019madex} \citep{shan2019research} \citep{kumari2010svm} \citep{yesmin2016identification} \citep{huang2004classifying}\\
        \multicolumn{3}{l}{URL: \url{https://www.csie.ntu.edu.tw/~cjlin/libsvm/}}\\
        \midrule
        Stanford CoreNLP toolkit\citep{manning2014stanford} &Tokenization, Lemmatization, POS tagging, NER, Word similarity& \citep{yang2019madex} \citep{Wang2018} \citep{dernoncourt-lee-2017-pubmed} \citep{li2020survey} \citep{li2017neural} \citep{tang2019detecting} \citep{filannino2018advancing} \citep{gu2016chemical} \citep{kilicoglu2020broad} \citep{dobreva2020improving} \citep{perera2020named} \citep{jofche2021pharmke} \citep{cunha2019recognizing} \citep{vzunic2020improving}\\
        \multicolumn{3}{l}{URL: \url{https://stanfordnlp.github.io/CoreNLP/}}\\
        \midrule
        BRAT\citep{stenetorp2012brat}  & \specialcell{Annotating structured text} &
        \citep{yang2021extracting} \citep{levitan2011application}\\
        \multicolumn{3}{l}{URL: \url{https://brat.nlplab.org/introduction.html}}\\
        \midrule
        SpaCy library\citep{spacy}  & Tokenization, Lemmatization, POS tagging, NER, Word similarity, SRL& \citep{peng2019transfer} \citep{li2020survey} \citep{jofche2021pharmke} \citep{mao2020use} \citep{dobreva2020improving} \citep{liu2019roberta} \citep{lai-2019-text-explanations} \citep{gururangan2020don} \citep{chen2020extracting} \citep{huang2019clinicalbert} \citep{rivera2019deep} \citep{d2021blockchain} \citep{tarcar2019healthcare} \citep{oyewusiartificial} \citep{zeng2022natural} \citep{jang2020exploratory} \citep{ramachandran2021named} \\
        \multicolumn{3}{l}{URL: \url{https://spacy.io/}}\\
        \midrule
        DOMEO\citep{ciccarese2012domeo} &  \specialcell{Annotating structured text}&
        \citep{hochheiser2016using} \citep{boyceusing}\\
        \multicolumn{3}{l}{URL: \url{https://github.com/domeo/domeo}}\\
        \midrule
        Transformers\citep{wolf-etal-2020-transformers} & NER, NLI, QA, SRL, Classification, Embeddings& \citep{kuratov2019adaptation} \citep{hussain2021pharmacovigilance} \citep{xiong-etal-2019-deep} \citep{beltagy-etal-2019-scibert} \citep{huang2019clinicalbert} \citep{el2020characterbert} \citep{Lee2019BioBERT} \citep{aldahdooh2021using} \citep{canete2020spanish} \citep{michalopoulos2021umlsbert} \citep{li2020effective} \citep{sun2021deep} \citep{Akhtyamova2020} \citep{peng2019transfer} \citep{dobreva2020improving} \citep{rogers2020primer} \citep{gururangan2020don} \citep{jofche2021pharmke} \citep{pfeiffer2021adapterfusion} \citep{breden2020detecting} \citep{houlsby2019parameter} \citep{khadhraoui2022survey} \citep{alsentzer-etal-2019-publicly} \citep{li2020survey} \citep{Sboev2022} \citep{mao2020use} \citep{lai-2019-text-explanations}\citep{moradi-2021-biotext-explanations} \citep{liu2021self} \citep{perera2020named} \citep{liu2019roberta} \citep{conneau2020unsupervised} \citep{aldahdooh2021r} \citep{yuan2022coder} \citep{qin2021entity} \\
        \multicolumn{3}{l}{URL: \url{https://huggingface.co/}}\\
        \midrule
        MedCat Tool\citep{kraljevic2019medcat}&NER+L& \citep{dobreva2022dd} \citep{alicante2016unsupervised}\\
         \multicolumn{3}{l}{URL: \url{https://github.com/CogStack/MedCAT}}\\
        \midrule
        AllenNLP\citep{gardner2018allennlp} & NER,NLI, QA, SRL, Classification, Embeddings&\citep{jofche2021pharmke} \citep{beltagy-etal-2019-scibert} \citep{Wang2018} \citep{dobreva2020improving} \citep{li2020survey} \citep{peng2019transfer} \citep{gururangan2020don} \citep{yang2021extracting} \citep{li2020lexicon}\\
        \multicolumn{3}{l}{URL: \url{https://allenai.org/allennlp}}\\
        \midrule
        Flair\citep{akbik2019flair}& \specialcell{NER, POS Tagging, Classification}&\citep{sun2021deep}\citep{Akhtyamova2020}\citep{conneau2020unsupervised}\\
        \multicolumn{3}{l}{URL: \url{https://github.com/flairNLP/flair}}\\
        \midrule
        Gensim\citep{rehurek_lrec} & \specialcell{Text summarization, Embeddings}&\citep{dobreva2020improving} \citep{Habibi2017deep} \citep{joshi2022unsupervised} \citep{dhrangadhariya2020machine} \citep{zhu2020doc2vec}\\
        \multicolumn{3}{l}{URL: \url{https://radimrehurek.com/gensim/}}\\
     
        \midrule
        JIEBA tool\citep{sun2012jieba}& Chinese words: POS tagging, TF-IDF, Text-Rank&        \citep{zhong2021internet} \citep{yangresearch} \citep{li2021hypergraph} \citep{yang2020extracting} \citep{lan2020research}\\
        \multicolumn{3}{l}{URL: \url{https://github.com/fxsjy/jieba}}\\
        \midrule
        TextBlob\citep{loria2018textblob} &NER, NLI, QA, SRL, Classification, Embeddings&
        \citep{sivasankari2017medical} \citep{saad2021determining} \citep{ribeiro2021discovering}\\
        \multicolumn{3}{l}{URL: \url{https://textblob.readthedocs.io/en/dev/}}\\
        \midrule
        Polyglot\citep{nystrom2003polyglot} & NER, POS Tagging, Sentiment Analysis, Embedding&\citep{li2020survey} \citep{prasad2013nextgen} \citep{ceusters2000language}\\
         \multicolumn{3}{l}{URL: \url{https://github.com/aboSamoor/polyglot}}\\
        \midrule
        Quepy\citep{andrawos2012quepy}& NLP, question transformation to queries&\citep{marginean2013towards}\\
        \multicolumn{3}{l}{URL: \url{https://github.com/machinalis/quepy}}\\
        \bottomrule
    \caption{Commonly used machine learning and NLP software libraries and tools.}
    \label{tab:libraries}
\end{longtable} 

\section{Conclusion}
\label{sec:conclusions}
Text is an important source of information in pharmacology. To extract that information from increasingly large collections of structured and unstructured documents, NLP is an essential approach. We present a survey of recent NLP developments relevant to the pharmacological domain. 

Our survey comprises five main pillars, each presented in its section: a modern methodology based on pretrained large language models, frequently used tasks, useful datasets, knowledge bases, and software libraries. 
Each main topic is further split into several components, giving our review a comprehensible hierarchical structure. We compress the main contributions of each section into overview tables at the end of each section. 
In summary, our survey testifies to swift developments in NLP and a surprising breadth of its use in pharmacology. 

While we reviewed over 250 works in our survey, the coverage is by no means exhaustive. In a few years, when next such a survey will be needed, we expect the most exciting developments in the use and integration of multi-modal resources, such as text, images, and 3D structural databases. In artificial intelligence, there is a tendency for large language models, called foundation models \citep{bommasani2021opportunities}, to capture as much human knowledge as possible, coupled with the ability for logical and commonsense reasoning. We expect that life sciences and pharmacology will be one of the first areas where domain-specific knowledge will be integrated into such models. 

Finally, NLP is a subfield of machine learning and artificial intelligence, which have many uses in pharmacology beyond NLP. We are not aware of any review comprehensively covering their applications in pharmacology, but such a work would complement ours. Due to broadness and rapid progress in ML and AI, such a review would require several research groups and a monograph format.

\section*{Acknowledgement}

This work is based on COST Action CA18209 – NexusLinguarum "European network for Web-centred linguistic data science", supported by COST (European Cooperation in Science and Technology). The work in this paper was partially financed by the Faculty of Computer Science and Engineering, Ss. Cyril and Methodius University in Skopje. The work was partially supported by the Slovenian Research Agency (ARRS) core research programme P6-0411 and the young researchers grant.

\bibliographystyle{plainnat} 
\bibliography{bibliography}

\end{document}